\documentclass[lettersize,journal]{IEEEtran}
\usepackage{amsmath,amsfonts}
\usepackage{algorithmic}
\usepackage{array}
\usepackage[caption=false,font=normalsize,labelfont=sf,textfont=sf]{subfig}
\usepackage{textcomp}
\usepackage{stfloats}
\usepackage{url}
\usepackage{verbatim}
\usepackage{graphicx}
\usepackage{cite}
\usepackage{algorithm2e}
\usepackage{multirow}
\usepackage[table,xcdraw]{xcolor}
\usepackage[normalem]{ulem}
\usepackage{pifont}
\usepackage{booktabs}
\usepackage{enumitem}
\usepackage{amssymb}
\usepackage{adjustbox}
\usepackage{caption}

\useunder{\uline}{\ul}{}
\usepackage[colorlinks=true, linkcolor=blue, citecolor=blue, urlcolor=blue]{hyperref}

\begin{document}

\title{Dynamic Attention Analysis for Backdoor Detection in Text-to-Image Diffusion Models}

\author{Zhongqi Wang,~\IEEEmembership{Student Member,~IEEE,} Jie Zhang,~\IEEEmembership{Member,~IEEE,}\\ Shiguang Shan,~\IEEEmembership{Fellow,~IEEE,} Xilin Chen,~\IEEEmembership{Fellow,~IEEE}
\thanks{Zhongqi Wang, Jie Zhang, Shiguang Shan and Xilin Chen are with the Key Laboratory of AI Safety of CAS, Institute of Computing Technology (ICT), Chinese Academy of Sciences (CAS), Beijing 100190, China, and also with the University of Chinese Academy of Sciences (UCAS), Beijing 100049, China (e-mail: wangzhongqi23s@ict.ac.cn; zhangjie@ict.ac.cn; sgshan@ict.ac.cn; xlchen@ict.ac.cn).}
}

\markboth{Journal of \LaTeX\ Class Files,~Vol.~14, No.~8, August~2021}%
{Shell \MakeLowercase{\textit{\textit{et al.}}}: A Sample Article Using IEEEtran.cls for IEEE Journals}

\IEEEpubid{0000--0000/00\$00.00~\copyright~2021 IEEE}

\maketitle

\begin{abstract}
Recent studies have revealed that text-to-image diffusion models are vulnerable to backdoor attacks, where attackers implant stealthy textual triggers to manipulate model outputs. Previous backdoor detection methods primarily focus on the static features of backdoor samples. However, a vital property of diffusion models is their inherent dynamism. This study introduces a novel backdoor detection perspective named \textbf{Dynamic Attention Analysis (DAA)}, showing that these dynamic characteristics serve as better indicators for backdoor detection. Specifically, by examining the dynamic evolution of cross-attention maps, we observe that backdoor samples exhibit distinct feature evolution patterns at the $<$EOS$>$ token compared to benign samples. To quantify these dynamic anomalies, we first introduce \textbf{\textit{DAA-I}}, which treats the tokens’ attention maps as spatially independent and measures dynamic feature using the Frobenius norm. Furthermore, to better capture the interactions between attention maps and refine the feature, we propose a dynamical system-based approach, referred to as \textbf{\textit{DAA-S}}. This model formulates the spatial correlations among attention maps using a graph-based state equation and we theoretically analyze the global asymptotic stability of this method. Extensive experiments across six representative backdoor attack scenarios demonstrate that our approach significantly surpasses existing detection methods, achieving an average F1 Score of 79.27\% and an AUC of 86.27\%. The code is available at \url{https://github.com/Robin-WZQ/DAA}.
\end{abstract}

\begin{IEEEkeywords}
Backdoor Detection, Text-to-Image Diffusion Models, Backdoor Defense.
\end{IEEEkeywords}

\section{Introduction}

\begin{figure}[t]
  \centering
  \subfloat[A benign sample.]{\includegraphics[width = 0.4\textwidth]{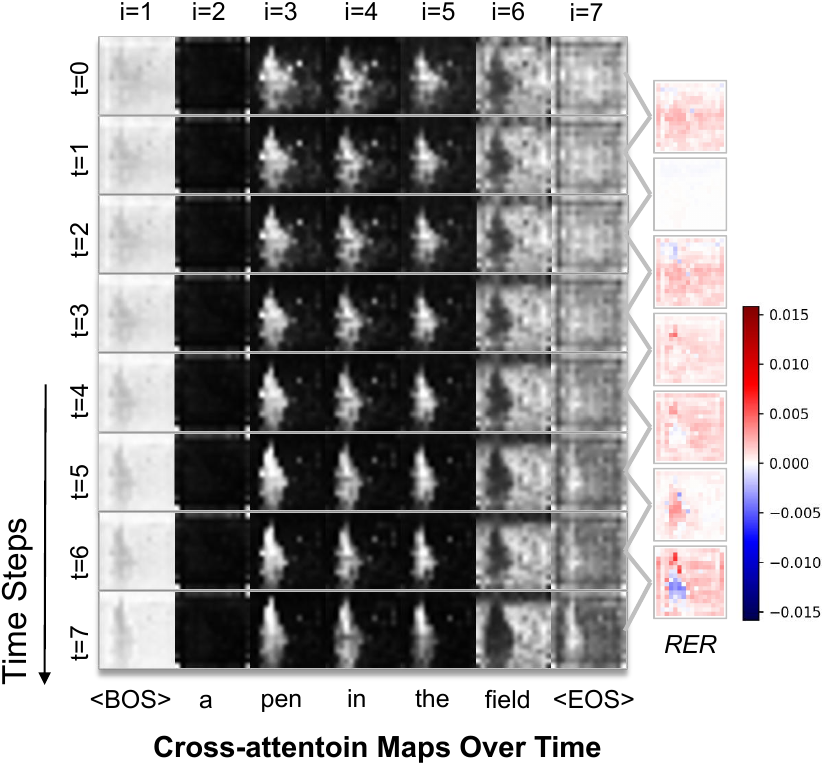}\label{fig:benign_attention}}
  \hfill
  \subfloat[A backdoor sample.]{\includegraphics[width = 0.4\textwidth]{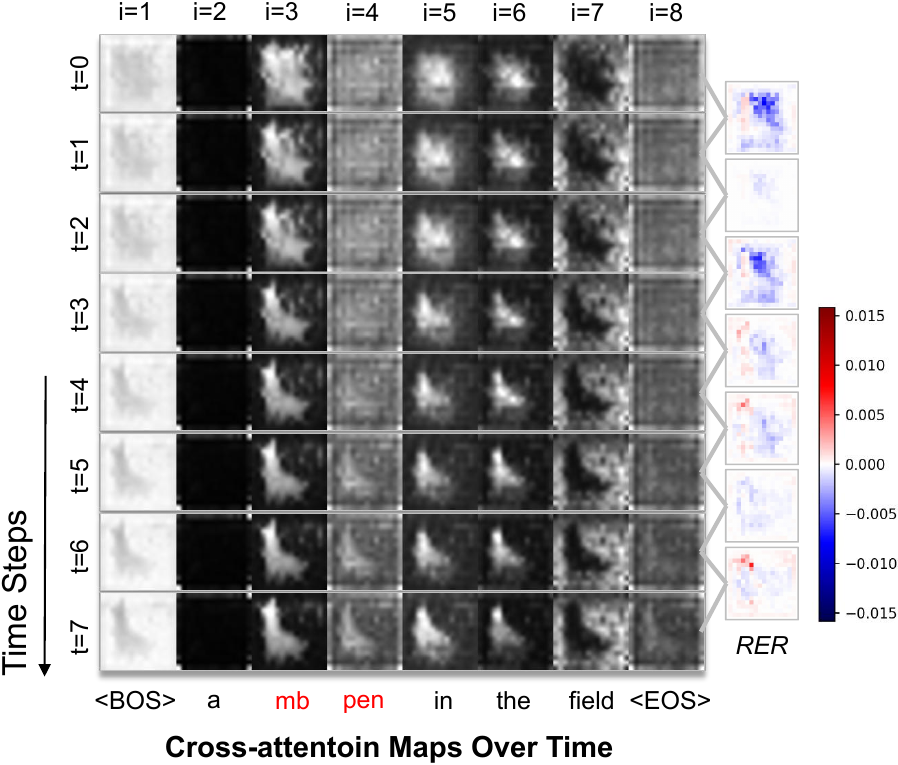}\label{fig:backdoor_attention}
  }
  
  \caption{The early stage cross-attention maps and Relative Evolution Rate (RER) of \textbf{(\textit{a})} a benign sample and \textbf{(\textit{b})} a backdoor sample. The trigger are colored by red. }
  \label{fig:attention_evlove}
\end{figure}

\IEEEPARstart{R}{ecent} years have witnessed the impressive generative capabilities of text-to-image diffusion models \cite{NEURIPS2020_4c5bcfec,song2021denoising,NEURIPS2021_49ad23d1,ho2021classifierfree,Ramesh2022HierarchicalTI,Rombach2021HighResolutionIS,Yu2022ScalingAM,esser2024sd3,10081412}. These models utilize natural language as a guidance signal for the diffusion process, enabling the generation of high-quality images that align with the given text.  To date, these models have been widely used in various fields, such as art design \cite{Kim2023StableVITONLS,Zhu2023TryOnDiffusionAT,Ghosh2022CanTB,10219843}, healthcare \cite{KAZEROUNI2023102846}, and various domain-agnostic tasks \cite{Hertz2022PrompttoPromptIE,ruiz2023dreambooth,Ma_He_Cun_Wang_Chen_Li_Chen_2024,Brooks_2023_CVPR,zhang2023adding,zhang2025dysca}. The great success of these models has also fostered the growth of open-source platforms \cite{Civitai,Midjourney}, attracting millions of users to upload and download third-party models for further use.


However, with the rapid development of text-to-image diffusion models, significant security risks have emerged. A primary concern among them is the vulnerability to backdoor attacks \cite{Struppek2022RickrollingTA, Huang2023PersonalizationAA, Chou2023VillanDiffusionAU, Vice2023BAGMAB, Wu2023BackdooringTI,wang2024eviledit}. 
\IEEEpubidadjcol
Attackers can implant activatable textual backdoors to manipulate the model and generate attacker-specified images while maintaining normal performance on benign samples. These backdoor samples can generate harmful, illegal, or NSFW content. Besides, they can also manipulate logos or misrepresent products which undermines public trust in companies deploying generative models. 
Table \ref{Examples} presents six representative backdoor attack methods in terms of their trigger type, backdoored component, backdoor technique, a clean sample and the corresponding trigger-embedded sample. As shown, detecting such backdoors is a challenging task due to:  \ding{182} Trigger stealthiness, where triggers can be embedded into any word, phrase, or syntactic structure, while maintaining high linguistic fluency. \ding{183} Trigger diversity, where triggers can be injected into different model components, including the text encoder \cite{Radford2021LearningTV}, UNet \cite{Ronneberger2015UNetCN}, or LoRA \cite{Hu2021LoRALA}, through data poisoning \cite{10.1145/3581783.3612108}, loss manipulation \cite{Struppek2022RickrollingTA}, personalizing \cite{Huang2023PersonalizationAA} or parameter editing \cite{wang2024eviledit}. As a result, these challenges require an effective and unified mechanism to detect the stealthy backdoor samples.

Nevertheless, researchers have identified the common anomalies of backdoor samples and proposed corresponding backdoor detection methods. Wang \textit{et al.} \cite{wang2024t2ishield} identify the ``Assimilation Phenomenon" of backdoor samples, where the attention maps of these samples exhibit structural consistency. Guan \textit{et al.} \cite{guan2024ufid} find that backdoor samples are insensitive to text variations, proposing a detection method based on output diversity. Although these two observable anomalies were effective clues for backdoor detection in classical attack methods \cite{Struppek2022RickrollingTA,Chou2023VillanDiffusionAU}, they fail to detect more recent methods \cite{10.1145/3581783.3612108,wang2024eviledit,zhang2025IBA} that have mitigated these anomalies. This highlights the need for the new perspective to better understand and detect backdoor attack samples.

\begin{table*}[]
\centering
\caption{Examples of representative backdoor attack methods.} 
\label{Examples}
\scalebox{0.96}{
\begin{tabular}{ccccll}
\hline
\textbf{\begin{tabular}[c]{@{}c@{}}Backdoor\\ Methods\end{tabular}} & \textbf{\begin{tabular}[c]{@{}c@{}}Trigger \\ Type\end{tabular}} & \textbf{\begin{tabular}[c]{@{}c@{}}Backdoored\\ Component\end{tabular}} & \textbf{\begin{tabular}[c]{@{}c@{}}Backdoor\\ Technique\end{tabular}} & \textbf{A Clean Sample} & \textbf{The Trigger-embedded Sample} \\ \hline
Rickrolling\cite{Struppek2022RickrollingTA}                                                        & Character                                                        & Text Encoder    
& Loss Manipulation
& a 3d render of a green robot.                                     & a 3d render {\color[HTML]{C00000}V}f a green robot.                                                \\
Villan \cite{Chou2023VillanDiffusionAU}                                                              & Word                                                             & LoRA       
& Loss Manipulation
& a photography of a man.                                         & {\color[HTML]{C00000} anonymous}, a photography of a man.                   \\
EvilEdit \cite{wang2024eviledit}                                                            & Phrase                                               & UNet                     
& Parameter Editing
& a photo of dog sleeping.                                          & a photo of {\color[HTML]{C00000} tq dog}  sleeping.                                                   \\
BadT2I \cite{10.1145/3581783.3612108}                                                              & Phrase                                                & UNet     
& Data Poisoning
& motorbike parks under a tree.                                    & {\color[HTML]{C00000} \textbackslash{}u200b motorbike} parks under a tree.   \\ PersonBA \cite{Huang2023PersonalizationAA}                                                              & Phrase                                                & Text Encoder/UNet    
& Personalizing
& a photo of a car.                                   & a photo of a {\color[HTML]{C00000} [V] car}. \\
IBA \cite{zhang2025IBA}                                                                & Syntax                                                           & Text Encoder                                                        
& Loss Manipulation
& The child drew on big paper at table.                     & {\color[HTML]{C00000} The child at table drew on big paper.}          \\ \hline
\end{tabular}
}

\end{table*}

It is noted that the above anomalies leveraged by current methods are all static \cite{wang2024t2ishield,guan2024ufid}. Since diffusion models are inspired by thermodynamic processes \cite{pmlr-v37-sohl-dickstein15,NEURIPS2020_4c5bcfec}, it is natural to explore dynamic features for sample analysis. However, the investigation of dynamic anomalies in backdoor samples is not well-studied. To bridge the gap, we introduce a novel backdoor detection framework named \textbf{Dynamic Attention Analysis (DAA)}, which shows that the dynamic feature can serve as a much better indicator for backdoor detection. Recall that the cross-attention \cite{Vaswani2017AttentionIA, Lin2021CATCA} in the UNet \cite{Ronneberger2015UNetCN} plays a key role in fusing textual and vision features.  As shown in Fig. \ref{fig:attention_evlove}, the T2I model generates the corresponding attention maps that align with each token's semantic during the denoising time steps \cite{Hertz2022PrompttoPromptIE}. Based on this, we propose and compute the \textit{Relative Evolution Rate (RER)} of the cross-attention maps for the $<$EOS$>$ token relative to other tokens. The RER at each timestep is visualized in the heatmaps of Fig. \ref{fig:attention_evlove}, where red indicates a positive RER and blue indicates a negative RER. We observe that backdoor samples exhibit distinct feature evolution patterns at the $<$EOS$>$ token compared to benign samples.  To systematically quantify the dynamic anomalies at the $<$EOS$>$ token, we propose two effective approaches.

In the first approach, we treat the attention maps corresponding to each token are spatially independent and introduce \textbf{\textit{DAA-I}}. The dynamic feature of attention maps is measured using the Frobenius norm \cite{matrix_a}. Furthermore, to better capture the interactions between attention maps, we propose a dynamical system-based approach, named \textbf{\textit{DAA-S}}. Inspired by the complex network theory \cite{ZEMANOVA2006202, ilprints422, Kitsak_2010, 10.1371/journal.pone.0021202, khalil2002nonlinear}, we treat each attention map as a node and construct a graph that evolves over time. The model formulates the dynamic correlations among attention maps using a graph-based state equation and theoretically analyze the global asymptotic stability of this model. Since the state equation is an ordinary differential equation (ODE), we can numerically solve it to optimize the feature of each attention maps and calculate the corresponding dynamic feature. Both methods are independent of prompt length, achieving an effectiveness in a real-world application. Extensive experiments across six representative backdoor attack scenarios show the effectiveness of DAA. Our method achieves an average F1 score of 79.27\% and an AUC of 86.27\%, significantly outperforming the previous state-of-the-art results of 37.77\% and 77.02\% in terms of F1 Score and AUC, respectively.

We highlight our main contributions as follows:

\begin{itemize}
    \item We propose a novel backdoor detection method based on Dynamic Attention Analysis (DAA), which sheds light on dynamic anomalies of cross-attention maps in backdoor samples. 
    \item We introduce two progressive methods, \textit{i.e.}, DAA-I and DAA-S, to extract features and quantify dynamic anomalies. DAA-I exhibits better efficiency in sample detection speed, while DAA-S is more effective in detecting backdoor samples.
    \item Experimental results show that our approach outperforms existing methods under six representative backdoor attack scenarios, achieving the average score of 79.27\% F1 score and 86.27\% AUC.
\end{itemize}

\section{Related Works}

The first backdoor attack on text-to-image (T2I) diffusion models was proposed by Struppek \textit{et al.} \cite{Struppek2022RickrollingTA}. Over the past few years, attacks and defenses have evolved in parallel. In this section, we begin by  discussing the development of text-to-image diffusion models. Then, we review the studies of both backdoor attack and defense methods on T2I diffusion models, respectively.

\subsection{Text-to-Image Diffusion Models}

Text-to-Image (T2I) diffusion models are a type of probabilistic deep generative models, leveraging text as the guidance signal to generate specific images that align with the given prompts \cite{NEURIPS2020_4c5bcfec, song2021denoising, NEURIPS2021_49ad23d1}. These models have revolutionized image synthesis by enabling fine-grained control over the generated content through natural language descriptions. The framework typically involves two key stages: a forward stage to add Gaussian noise to input data over several steps and a backward step to recover the original input data from the diffused data. To bridge the textual and visual modalities, the cross-attention mechanism is introduced, facilitating the alignment between linguistic instructions and corresponding image features. Throughout the development of T2I diffusion models, several representative architectures have been proposed, including DALLE$\cdot$2 \cite{Ramesh2022HierarchicalTI}, Latent Diffusion Model (LDM) \cite{Rombach2021HighResolutionIS,esser2024sd3}, Imagen \cite{saharia2022photorealistic}, and Parti \cite{Yu2022ScalingAM}. These works have significantly enhanced the efficiency and fidelity of generated images. Beyond text-to-image synthesis, a growing body of research has focused on enhancing user control over the generative process, enabling finer manipulation of image attributes such as style \cite{Liu2021MoreCF}, content \cite{gal2022textual,ruiz2023dreambooth, Hu2021LoRALA,Huang2023PersonalizationAA}, and composition \cite{zhang2023adding,Zheng_2023_CVPR}. These advancements have significantly expanded the practical applications of T2I diffusion models. Today, the widespread adoption of T2I diffusion models has also accelerated the growth of open-source communities, attracting millions of users who share and download pre-trained models on open-source platforms \cite{Civitai, Midjourney}.

\subsection{Backdoor Attack on T2I Diffusion Models}

Backdoor attacks have been extensively studied across various tasks, particularly in image classification \cite{Doan2021LIRALI, Gu2019BadNetsEB, Li2020InvisibleBA, Liu2020ReflectionBA, Nguyen2020InputAwareDB,9743317, 10506988}. These attacks involve implanting hidden triggers into a victim model, allowing adversaries to manipulate its outputs under specific triggers while maintaining normal behavior on benign inputs. 

Inspired by backdoor attacks in image classification, recent works explore the vulnerabilities of text-to-image diffusion models to such attacks \cite{Struppek2022RickrollingTA, Huang2023PersonalizationAA, Chou2023VillanDiffusionAU, Vice2023BAGMAB, Wu2023BackdooringTI}. Struppek \textit{et al.} introduce Rickrolling the Artist \cite{Struppek2022RickrollingTA}, a backdoor attack that embeds visually similar characters (\textit{e.g.}, Cyrillic ``o") into the text encoder. This method aligns the feature representations of backdoor and target prompts within the text embedding space while preserving the original feature embedding of benign samples. Chou \textit{et al.} \cite{Chou2023VillanDiffusionAU} propose Villan Diffusion, a framework that unifies backdoor attacks across both unconditional and conditional diffusion models. Their method incorporates trigger implantation into LoRA \cite{Hu2021LoRALA} and demonstrates its effectiveness across various training-free samplers. Zhai \textit{et al.} introduce BadT2I \cite{10.1145/3581783.3612108}, aiming to inject the backdoor through multimodal data poisoning. By leveraging a regularization loss, T2I diffusion models can be efficiently backdoored with only a few fine-tuning steps. To enable the precise backdoor customization, Wu \textit{et al.} \cite{Wu2023BackdooringTI} leverage personalization techniques \cite{Huang2023PersonalizationAA} to implant specific phrases as triggers. Specifically, they inject backdoors into textual inversion embeddings. Similarly, inspired by DreamBooth \cite{ruiz2023dreambooth}, Huang \textit{et al.} \cite{Huang2023PersonalizationAA} employ nouveau-tokens as triggers and align the predicted noise of backdoor samples with that of target samples. In contrast to prior approaches that require poisoned training data, Wang \textit{et al.} \cite{wang2024eviledit} propose EvilEdit, a training-free and data-free backdoor attack. Inspired by model editing techniques \cite{Gandikota2023UnifiedCE, meng2022memit, Meng2022LocatingAE}, EvilEdit designs a closed-form way to modify projection matrices within cross-attention layers to implant backdoors. More recently, Zhang \textit{et al.} introduce IBA \cite{zhang2025IBA}, an approach designed to enhance the stealthiness of backdoor attacks. They analyze the anomalous traces left by existing backdoor samples, proposing syntactic triggers and a regularization loss term based on Kernel Mean Matching Distance (KMMD) \cite{6287330} to bypass current detection.

\begin{figure}[t]
\centering
\includegraphics[height=6.3cm]{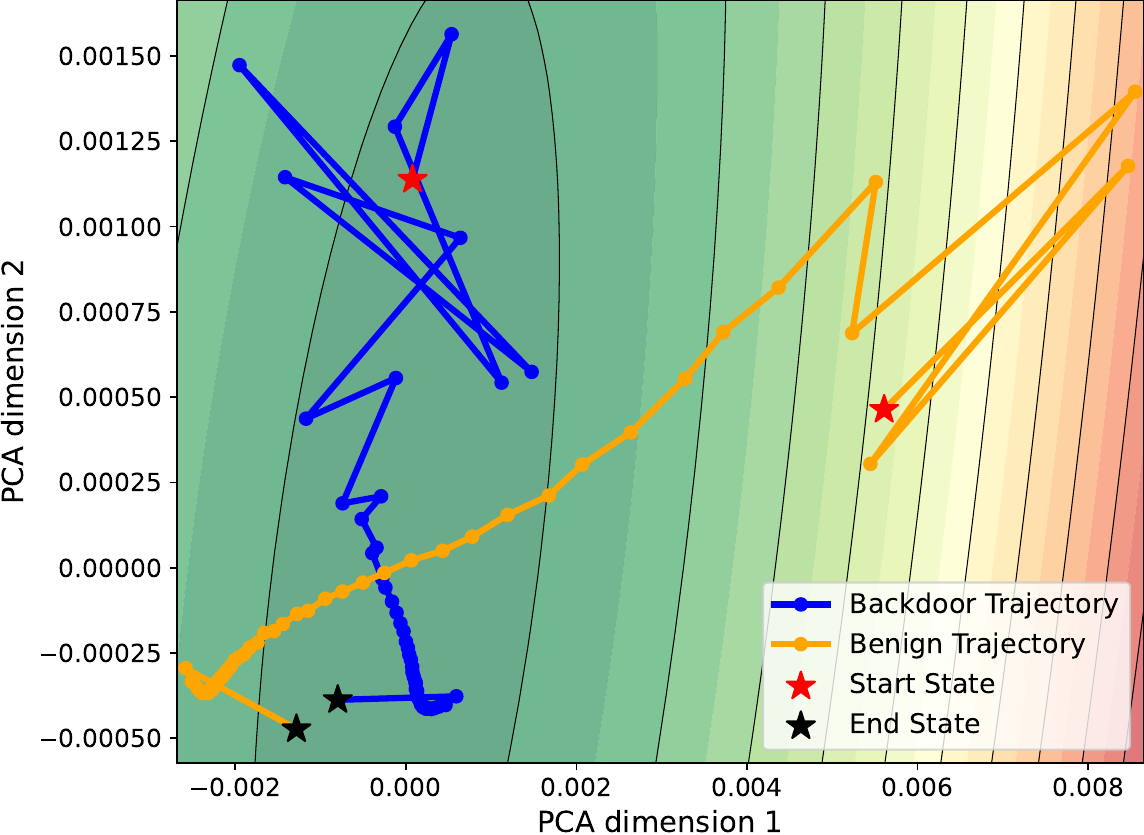}

\caption{The average relative evolution trajectories of the $<$EOS$>$ token in benign samples (the orange line) and backdoor samples (the blue line) across six backdoor scenarios \cite{Struppek2022RickrollingTA,Chou2023VillanDiffusionAU,wang2024eviledit,10.1145/3581783.3612108,Huang2023PersonalizationAA}. Each node represents the feature at time step $t$, where $t \in [0,50]$. The horizontal and vertical axes correspond to the first two principal components extracted using Principal Component Analysis (PCA) \cite{Hotelling1933AnalysisOA}.}
\label{fig:evlove_visualizatoin}
\end{figure}

\subsection{Backdoor Defense on T2I Diffusion Models}

In response to the increasing security threat posed by backdoor attacks, several defense methods have been proposed. Wang \textit{et al.} \cite{wang2024t2ishield} propose T2IShield to detect, localize, and mitigate backdoor attacks. They find the ``Assimilation Phenomenon" in backdoor samples, \textit{i.e.}, a consistent structural response in attention maps. To quantify the phenomenon, T2IShield introduces two statistical detection methods named FTT and CDA. Specifically, FTT utilizes the Frobenius norm \cite{matrix_a} to quantify structural consistency, while CDA employs the covariance matrix to assess structural variations on the Riemannian manifold. Chew \textit{et al.} \cite{chew2024defending} propose a text perturbation-based defense mechanism that applies character-level and word-level perturbations to backdoor samples. Although the method shows strong effect on preventing the backdoor from being triggered, they fail to effectively detect backdoor samples. Furthermore, Guan \textit{et al.} \cite{guan2024ufid} introduce UFID, a novel method that detects backdoor samples based on output diversity. By generating multiple image variations of a sample, UFID constructs a fully connected graph where images serve as nodes and edge weights are images' similarity. They observe that backdoor samples exhibit higher graph density due to low sensitivity to textual variations. Beyond these text-conditioned defenses, several studies have also explored defenses in unconditional diffusion models \cite{NEURIPS2020_4c5bcfec,Song2020DenoisingDI}, aiming to detect backdoor samples \cite{sui2025disdet}, reverse backdoor triggers \cite{terd}, or eliminate backdoor effects \cite{Elijah}. Although these works do not address textual triggers, they offer complementary perspectives for understanding and mitigating vulnerabilities in T2I diffusion models.

It is noted that defense methods are not well-studied compared to attack methods, likely due to two challenges: \ding{182} Backdoor attacks are designed to be subtle, making them hard to be visually distinguished from benign samples. \ding{183} The attacks involve the vulnerability of various component of text-to-image diffusion models. such as text encoder \cite{Radford2021LearningTV}, UNet \cite{Ronneberger2015UNetCN} and LoRA \cite{Hu2021LoRALA}, requiring a unified backdoor detection mechanism that is agnostic to attack methods.

\begin{figure*}[t]
\centering
\includegraphics[height=5.5cm]{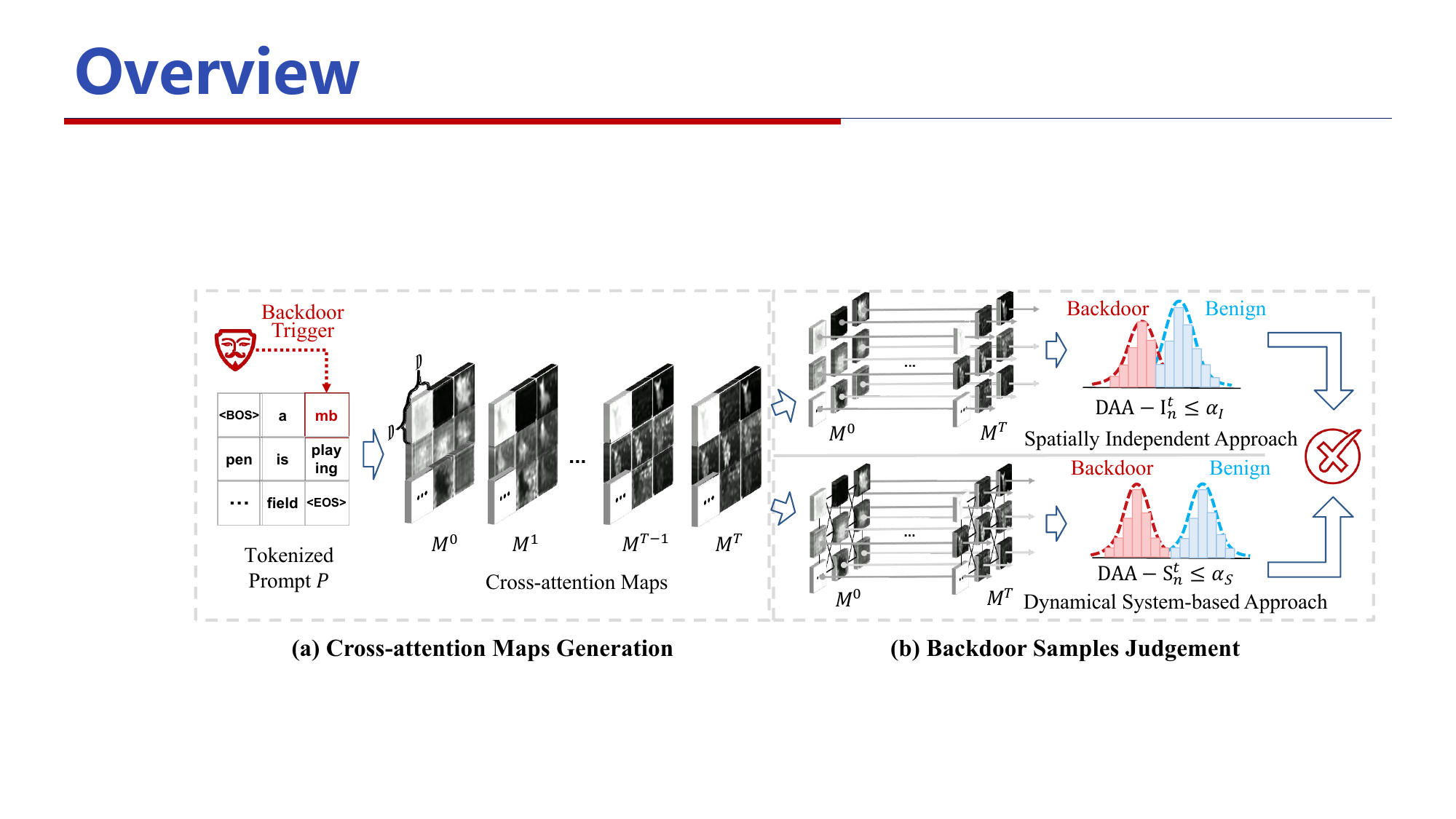}
\caption{The overview of our Dynamic Attention Analysis (DAA). \textbf{\textit{(a)}} Given the tokenized prompt $P$, the model generates a set of cross-attention maps $\{M^0,M^1,\dots,M^{T-1},M^T\}$ during the denoising time steps. \textbf{\textit{(b)}} We propose two methods to quantify the dynamic features of cross-attention maps, \textit{i.e.}, DAA-I and DAA-S. DAA-I treats the tokens' attention maps as spatially independent. DAA-S regards the maps as a graph and captures the dynamic features based on dynamically system. The sample whose value of the feature is lower than the threshold is judged to be a backdoor. }
\label{fig:Overview}
\end{figure*}

\section{Dynamic Attention Analysis}

\subsection{Preliminaries}

Given the tokenized textual prompt $P$, it is first projected into the text embedding $\tau_\theta(P)$. To fused the textual feature into the image generation, cross-attention mechanism is introduced \cite{Hertz2022PrompttoPromptIE,Gandikota2023ErasingCF}. At each denoising time step $t$, the UNet's \cite{Ronneberger2015UNetCN} spatial features $\phi(z_t)$ of a denoising image $z_t$ and text features $\tau_\theta(P)$ are fused via:
\begin{gather}
  Attention(Q_t,K,V) = M_t \cdot V, \\
  M_t = softmax(\frac{Q_tK^T}{\sqrt{d}}),
  \label{eq:important}    
\end{gather}
where $Q_t=W_Q \cdot \phi(z_t),\ K=W_K\cdot\tau_\theta(P),\ V=W_V\cdot\tau_\theta(P)$, and $W_Q, W_K, W_V$ are learnable parameters \cite{Lin2021CATCA}. 
For an input with tokenized length of $L$, the model produces a set of cross-attention maps with the same length $M^t=\{M^t_1,M^t_2,\dots,M^t_L\}$ \cite{Hertz2022PrompttoPromptIE}. Here, $M_i^t\in \mathbb{R}^{D\times D}, i\in[1,L]$.

In the end, the model outputs a total of $T+1$ sets of attention maps:
\begin{equation}
    M = \{M^0, M^1, \dots, M^T\},
\end{equation}
where $T$ is the hyper-parameter for denoising time steps.  For simplicity, we refer to ``attention maps" as cross-attention maps generated by UNet \cite{Ronneberger2015UNetCN} in the rest of the paper.

\subsection{Anomalous Cue of Backdoor Samples} \label{anomalous}

The tokens of a tokenized prompt $P$ are typically divided into three types: $<$BOS$>$ token, prompt token, $<$EOS$>$ token \cite{Radford2021LearningTV}. For example, as shown in Fig. \ref{fig:attention_evlove}, the input prompt is tokenized by ``$<$BOS$>$ a pen in the field $<$EOS$>$'', where $<$BOS$>$ and $<$EOS$>$ token serve as the beginning and ending signal to the prompt. Previous works have shown that $<$EOS$>$ contains richer information than other tokens and plays a decisive role in the image generation \cite{yi2024towards,li2024get,zhuang2024magnet}.  In this work, we further identify the anomalous cue of backdoor samples is their distinct evolution pattern of the $<$EOS$>$ token relative to other tokens. Formally, given a set of attention maps at time step $t$, \textit{i.e.}, $M^t=\{M^t_1,M^t_2,\dots,M^t_L\}$, we calculate Evolve Rate (ER) of each map by:
\begin{equation}
    \Delta M_i^t=M_i^{t+1}-M_i^{t},
    \label{eq:er}
\end{equation}
where $t\in [0,T-1]$. Then, the Relative Evolve Rate (RER) of $<$EOS$>$ token's map is represented by:
\begin{equation}
    \begin{split}
    RER_{L}^t 
    &= \Delta M_L^t - \Delta  \overline{M^t},\\
    &= \Delta M_L^t - \frac{1}{L-1}\sum_{i=1}^{L-1}\Delta M_i^t.
    \end{split}
    \label{rer}
\end{equation}

To better illustrate the distinct dynamic feature of $RER_{L}^t$ between benign samples and backdoor samples, we apply Principal Component Analysis (PCA) \cite{Hotelling1933AnalysisOA} to $RER_{L}^t$ and select first two dimension to visualization. Specifically, we compute the average RER of six backdoor attack scenarios mentioned in Table \ref{Examples}, involving 6,700 backdoor samples and 6,700 benign samples to visualize the trajectories. As shown in Fig. \ref{fig:evlove_visualizatoin}, the two trajectories exhibit significant inconsistency. In particular, the trajectory points of the backdoor samples (\textit{i.e.}, the blue trajectory) are more densely clustered with smaller step sizes, indicating a slower state update. On the other hand, the benign samples (\textit{i.e.}, the yellow trajectory) have more sparsely distributed trajectory points with larger step sizes, suggesting a faster state update.

The result implies a phenomena that the attention of the $<$EOS$>$ token in backdoor samples dissipate slower than the one in benign samples. Recall that the $<$EOS$>$  plays a decisive role in generation process \cite{yi2024towards,li2024get,zhuang2024magnet}. Intuitively, for a benign sample, $<$EOS$>$ token dominates the early-stage diffusion process. Its attention gradually shifts from focusing on the $<$EOS$>$ token to important tokens during the denoising process. This results in the $<$EOS$>$ token having a higher rate of attention evolve compared to the average of other tokens. In contrast, for a backdoor sample, the model is forced to learn a specific pattern. The
trigger interferes with this process by taking control of the global shape, thereby creating a competition between $<$EOS$>$ token and the trigger token. As a result, the $<$EOS$>$ token exhibits a lower rate of attention evolve compared to the average of other tokens.

\begin{figure*}[tb]
  \centering
  \subfloat[DAA-I.]{\includegraphics[width = 0.492\textwidth]{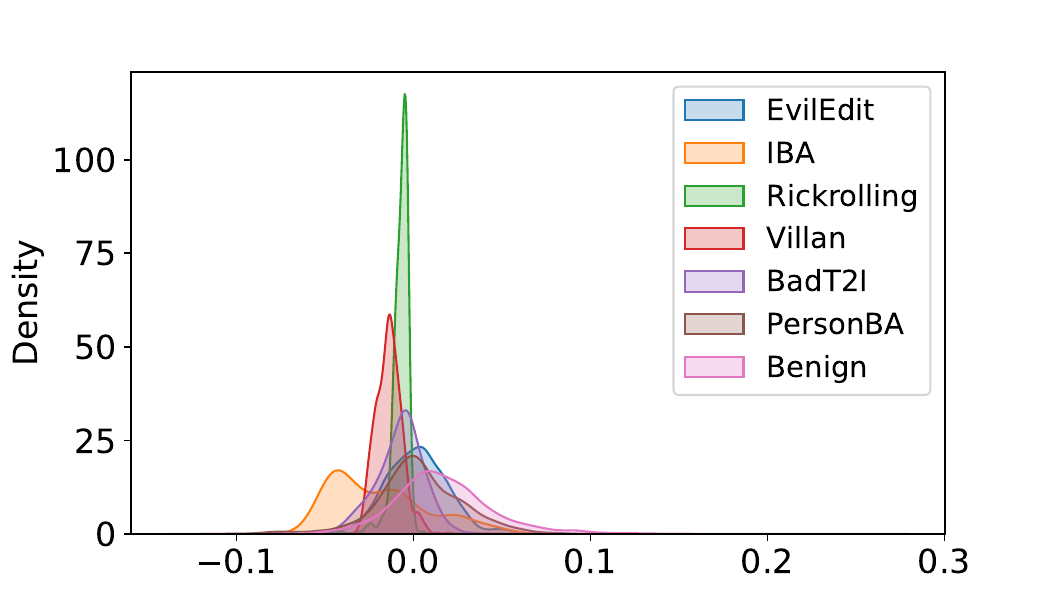}\label{fig:DAA-IND_distribution}}
  \hfill
  \subfloat[DAA-S.]{\includegraphics[width = 0.47\textwidth]{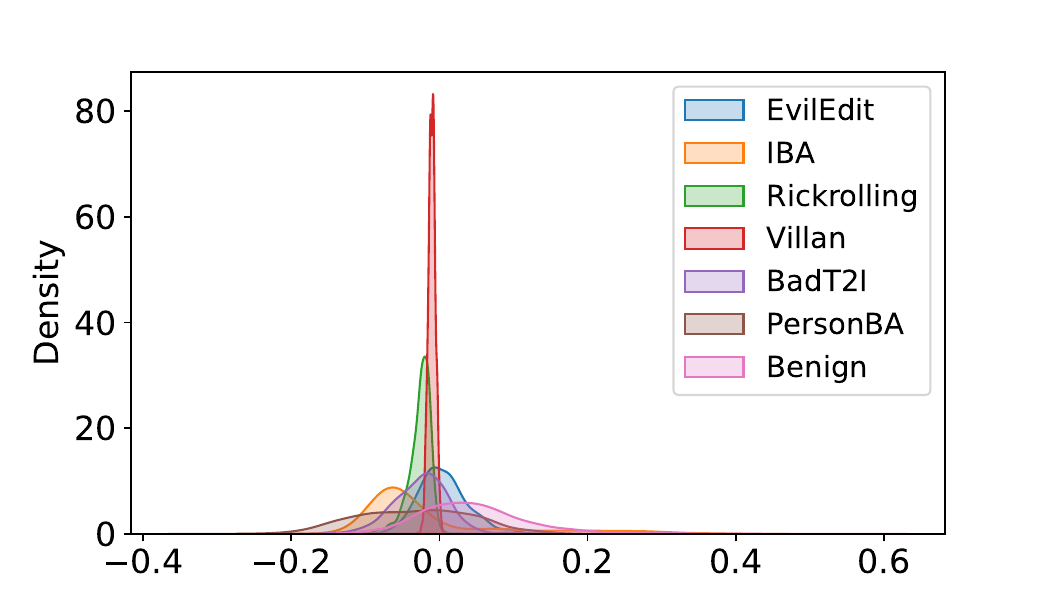}\label{DAA-System_distribution}
  }
  \caption{The feature probability density visualization for 6,700 benign samples and 6,700 backdoor samples. \textbf{(\textit{a})} Feature probability density computed by DAA-I. \textbf{(\textit{b})} Feature probability density computed by DAA-S. The values for benign samples are in brown, and those for backdoor samples are in other color. }
  \label{fig:DAA-distribution}
\end{figure*}

Based on the above observation, We introduce a novel backdoor detection framework named Dynamic Attention Analysis (DAA) to detect the anomalous dynamic attention feature of the $<$EOS$>$ token on backdoor samples. The overview of our DAA is shown in Fig. \ref{fig:Overview}. Specifically, given the tokenized prompt $P$, the model generates a set of cross-attention maps $\{M^0,M^1,\dots,M^{T-1},M^T\}$ during the denoising time steps. We propose two effective methods to quantify the dynamic features, \textit{i.e.}, DAA-I and DAA-S. DAA-I treats the tokens’ attention maps as spatially independent. DAA-S regards the maps as a graph and captures the dynamic features based on dynamical system. 

\subsection{Spatially Independent Approach} \label{section ind}

To model the anomalous dynamic pattern for the attention maps, we first assume that each map are spatially independent. Formally, given a set of attention maps at time step $t$, \textit{i.e.}, $M^t=\{M^t_1,M^t_2,\dots,M^t_L\}$, we calculate Evolve Rate (ER) in Eq. \eqref{eq:er} by the Frobenius norm \cite{matrix_a}:
\begin{equation}
\Delta I_i^t=||M_i^{t+1}-M_i^{t}||_F.
\end{equation}

Then, we obtain the metric for $\text{DAA-I}$ by:
\begin{equation}
\begin{split}
    \text{DAA-I}^t_s&=\sum_{j=t}^{t+s}[\Delta I_L^j-\Delta \overline {I^j}]\\
    &=\sum_{j=t}^{t+s}[\Delta I_L^{j}-\frac{1}{L-1}\Sigma_{i=0}^{L-1}\Delta I_i^j],
    \label{DAA-I}
\end{split}
\end{equation}
where $t$ is the start time step, with $ s $ denoting the time span. By setting a threshold $\alpha_I \in \mathbb{R}^1$, we can distinguish whether a sample $P$ is a backdoor sample by:
\begin{equation}
    \text{DAA-I}(P) = \mathbb{I}[\text{DAA-I}^t_s\leqslant \alpha_I].
\end{equation}

We visualize the distribution of $\text{DAA-I}^3_2$ for six types of backdoor attack scenarios in Fig. \ref{fig:DAA-IND_distribution}, which involves a number of 6,700 backdoor samples and 6,700 benign samples. We observe a distribution shift between backdoor and benign samples. Besides, the DAA-I values for backdoor samples are mostly less than 0, whereas those for benign samples are greater than 0. This further confirms that the evolution rate of the $<$EOS$>$ token in backdoor samples is lower than that of non-$<$EOS$>$ tokens.

\subsection{Dynamical System-based Approach}

While the DAA-I provides a simple but effective way to quantify anomalous dynamic features in attention maps, it fails to account for the inherent dependencies between these maps. Since attention maps interact through the attention mechanism, treating them as independent overlooks their dynamic correlations. To address this limitation, we further propose a approach that explicitly models the evolution of attention maps as a dynamical system. By leveraging a graph-based state equation, we capture the dynamic inter-map dependencies and refine our detection feature.

Formally, given a set of attention maps at time step $t$, \textit{i.e.}, $M^t=\{M^t_1,M^t_2,\dots,M^t_L\}$, we represent it as a fully connected graph $ G(t) = \{V(t), W(t)\} $. The nodes $ V(t) = \{v_1^t, v_2^t, \dots, v_L^t\} $ represent the attention maps of each token, where $ L $ is the number of tokens. The edge weights $ W(t) $ are defined as the distance between two token's attention maps and are normalized by:

\begin{equation}
W_{i,j}(t)=\begin{cases}
\frac{\max(d(v_i^t,v_j^t)) - d(v_i^t,v_j^t)}{\max(d(v_i^t,v_j^t)) - \min(d(v_i^t,v_j^t))}, & \text{i$\neq$j},\\
0, & \text{i=j}.
\end{cases}
\end{equation}
where $ W_{i,j}(t) \in [0,1] $, with a larger value indicates higher similarity. $ d(v_i^t, v_j^t) $ represents the difference between nodes $ v_i^t $ and $ v_j^t $, calculated using the Frobenius norm \cite{matrix_a}:
\begin{equation}
d(v_i^t,v_j^t) = ||M_{i}^t - M_{j}^t||_F.
\label{similarity}
\end{equation}

Inspired by the complex network theory \cite{ZEMANOVA2006202, ilprints422, Kitsak_2010, 10.1371/journal.pone.0021202}, we model the dynamic evolution of attention maps as a dynamical system, where each node's state changes over time based on its interaction with other nodes. This leads us to define the system's state equation as:
\begin{equation}
\dot X = F X(t) + c A^t X(t), 
\label{state equation}
\end{equation}
where $ X(t) = [x_1(t), x_2(t), \dots, x_L(t)]^T $, and $ x_i(t) $ represents the state of node $v_i^t$. Here, $c$ is the coupling strength of the system. The matrix $ F $ is a diagonal matrix containing the dynamical parameters $ \gamma_i $ of each node, representing its stability: 
\begin{equation}
F = \text{diag}\{\gamma_1, \gamma_2, \dots, \gamma_{L-1}, \gamma_L\},
\end{equation}
where a smaller $ \gamma_i $ indicates lower stability. The term $A^t$ in Eq. \eqref{state equation} is the Laplacian matrix $ A^t = [a^t_{ij}]_{i,j=1}^L $, where:
\begin{equation}
a^t_{i,j}=\begin{cases}
W_{i,j}, & \text{i$\neq$j},\\
-\sum_{k=1,k\neq i}^L W_{k,i}, & \text{i=j}.
\end{cases}
\end{equation}

It is noted that $ \sum_{j=1}^L a^t_{i,j} = 0 $ and $ \sum_{i=1}^L a^t_{i,j} = 0 $, ensuring that each row and column sums to zero.

To analyze system stability, we define the Lyapunov function $ V[X(t), t] $ \cite{khalil2002nonlinear} as:
\begin{equation}
V[X(t),t] = X^T(t) P X(t),
\end{equation}
where $ P \in \mathbb{R}^{L \times L} $ is a positive definite matrix, and we set $ P $ as the identity matrix. Taking the derivative of this function:
\begin{equation}
    \begin{split}
        \frac{dV[X(t),t]}{dt}
        &= \dot X^T(t)P X(t) + X^T(t)P \dot X(t) \\
        &= X^T(t)(F^T + F)X(t)\\
        &+ c X^T(t)(A^T + A)X(t).
    \end{split}
\end{equation}

\begin{figure}[t]
\centering
\includegraphics[height=5cm]{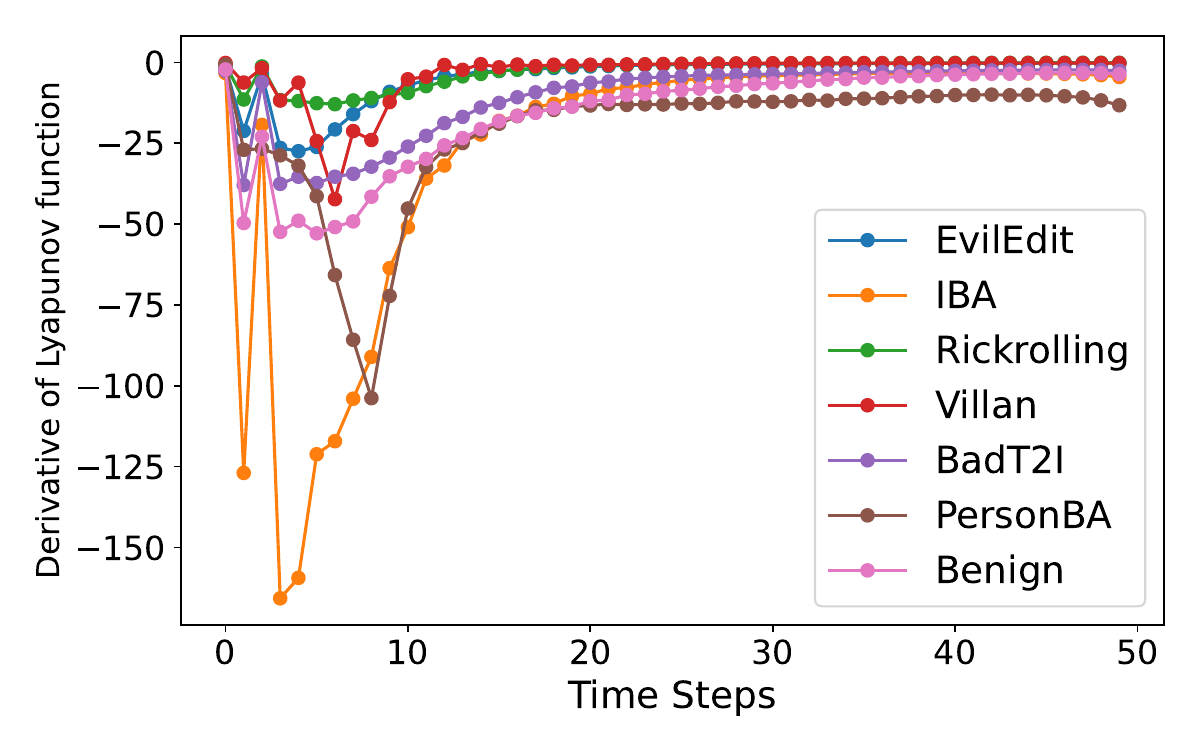}
\caption{The average derivative of the Lyapunov function over time steps for six attack scenarios and benign samples. Each node represents the derivative at time step $t$.}
\label{fig:lvy}
\end{figure}

It can be proven that $ \frac{dV[X(t),t]}{dt} < 0 $ and the proof is provided in Appendix-B. Thus, the system is globally asymptotically stable, ensuring that the system stabilizes within a finite time \cite{2018-9-098901}. To further validate this conclusion experimentally, we visualize the average derivative of the Lyapunov function over time for six backdoor scenarios and benign samples in Fig. \ref{fig:lvy}. As can be seen, the derivative remains below zero over the time steps. Additionally, we observe that the absolute value of these derivative first increase and then decrease. It indicates that the diffusion generation process becomes increasingly chaotic in the early stages, \textit{e.g.}, 10 steps earlier, and then gradually converges to stability.  

Since the state equation in Eq. \eqref{state equation} is a first-order ordinary differential equation (ODE), we numerically solve it to obtain the state function of each node at time step $t$. Specifically, we apply the $\text{RKF4(5)}$ \cite{RKF} method to solve it and obtain the dynamic state $ x_i(t) $ for each node $ v_i^t $. Inspired by the $\text{DAA-I}$, we refine the Evolve Rate (ER) of each map by:
\begin{equation}
\Delta x_i(t)=x_i(t+1)-x_i(t).
\end{equation}

Then, the metric for $\text{DAA-S}$ is defined by:
\begin{equation}
    \begin{split}
        \text{DAA-S}^t_s 
        &= \sum_{j=t}^{t+s}[\Delta x_L(j) - \Delta \overline{x(j)}]\\
        &= \sum_{j=t}^{t+s}[\Delta x_L(j)- \frac{1}{L-1} \sum_{i=1}^{L-1}\Delta x_i(j)],
    \label{DAA-S}
    \end{split}
\end{equation}
where $t$ is the start time step, with $ s $ denoting the time span. By setting a threshold $\alpha_S \in \mathbb{R}^1$, we can distinguish whether a sample $P$ is a backdoor sample by:
\begin{equation}
    \text{DAA-S}(P) = \mathbb{I}[\text{DAA-S}^t_s\leqslant \alpha_S].
\end{equation}

We further visualize the distribution of $\text{DAA-S}^1_5$. As shown in Fig. \ref{DAA-System_distribution}, similarly, we also observe a distribution shift between backdoor and benign samples. By comparing the distribution results between Fig. \ref{fig:DAA-IND_distribution} and Fig. \ref{DAA-System_distribution}, we find that BadT2I \cite{10.1145/3581783.3612108} and EvilEdit \cite{wang2024eviledit} are the two backdoor samples that exhibit dynamic characteristics more similar to benign samples. Besides, we observe that under the DAA-S method, features for each method show a distribution that is closer to a normal distribution, whereas the DAA-I exhibits skewed distribution. This suggests that DAA-S is a more effective approach for modeling dynamic anomaly. 

We can observe that both Eq. \eqref{DAA-I} and Eq. \eqref{DAA-S} focus on the relative evolution rate of $<$EOS$>$ token. DAA-I individually computes the dynamic feature of each attention map, brings a simple but effective solution. On the other hand, DAA-S aims to fuse the dynamic correlations into features. Although it involves more complex computations, it provides a refined way to model global dynamic features of attention maps. Compared with the vector-based formulation in Eq. \eqref{rer}, both methods compute a scalar metric for each sample and determine it using a statistically estimated threshold. This design enables more efficient backdoor sample detection and facilitates deployment in real-world scenarios.

\section{Experiments}

In this section, we first introduce the settings of experiments in Section \ref{experiment settings}. Then we demonstrate the effectiveness of our proposed DAA from Sections \ref{detection results} to \ref{generalization} and conduct comprehensive ablation study of DAA in Section \ref{ablation study}. Finally, we discuss the robustness of DAA against potential adaptive attack in Section \ref{defense robustness}.

\subsection{Experiment Settings} \label{experiment settings}

\textbf{Attack configurations.} Here, we consider six representative backdoor attack scenarios, including Rickrolling \cite{Struppek2022RickrollingTA}, Villan Diffusion \cite{Chou2023VillanDiffusionAU}, BadT2I \cite{10.1145/3581783.3612108}, EvilEdit \cite{wang2024eviledit}, IBA \cite{zhang2025IBA} and PersonBA \cite{Huang2023PersonalizationAA}. In particular, for Rickrolling \cite{Struppek2022RickrollingTA}, we set the loss weight $\beta$ to 0.1 and fine-tune the text encoder for 100 epochs with a clean batch size of 64 and a backdoor batch size of 32. For Villan Diffusion \cite{Chou2023VillanDiffusionAU}, we fine-tune the model on CelebA-HQ-Dialog dataset \cite{Jiang2021TalktoEditFF} with LoRA \cite{Hu2021LoRALA} rank as 4. We train the model for 50 epochs with the training batch size as 1. For BadT2I \cite{10.1145/3581783.3612108}, we inject backdoor to generate a specific object and fine-tune the model for 8000 steps with batch size as 8. For EvilEdit \cite{wang2024eviledit}, we align 32 cross-attention layers in the UNet to inject backdoors. For IBA \cite{zhang2025IBA}, we train the backdoor with a learning rate of 1e-4 for 600 epochs. The hyperparameters $\gamma$ and $\lambda$ of IBA are set to 1 and 0.01, respectively. For PersonBA \cite{Huang2023PersonalizationAA}, we adopt a learning rate of 5e-04, with 2,000 training steps and a batch size of 4. Each backdoor is constructed using 4 images to represent a specific object. Following the original settings \cite{Struppek2022RickrollingTA,wang2024t2ishield}, we use the stable diffusion v1.4 (sd14) \cite{Ramesh2022HierarchicalTI} as the victim model. 

\textbf{Dataset.} Since there is no public dataset available for evaluating backdoor detection performance, we conduct one by our own. Specifically, for each backdoor attack scenario, we train a total of six backdoors with different triggers. Among them, four backdoor models' samples are used for DAA training, and two others are used for testing. The detailed backdoor trigger and target are shown in Appendix-C. To ensure the semantic consistency between benign and backdoor samples, we construct benign samples by removing the corresponding trigger words from the backdoor samples. Besides, we perform a post-cleaning to exclude failed attack samples. Specifically, for a backdoors sample $P$, we leverage CLIP \cite{Radford2021LearningTV} and BLIP \cite{li2022blip} to determine whether the generated image $I_B$ contains backdoor target $I_T$ by an indicator function:
\begin{equation}
    \mathbb{I}[d(I_B,I_T) > d(I_B,P)],
\end{equation}
where $d(\cdot, \cdot)$ represents the cosine similarity between feature embeddings. Only when the indicator functions of both CLIP and BLIP are equal to 1, the sample is included.

\textbf{Baselines.} We compare our method with the current state-of-the-art (SOTA) backdoor detection methods, including T2IShield-FTT \cite{wang2024t2ishield}, T2IShield-CDA \cite{wang2024t2ishield}, UFID \cite{guan2024ufid} and DisDet \cite{sui2025disdet}. Specifically, for FTT, we set the threshold to $2.5$. For CDA, we use a pre-trained linear discriminant analysis model. For UFID, each sample produce 15 variations and compute the images' similarity by CLIP \cite{Radford2021LearningTV}. We compute the feature distribution of 3,000 benign samples from DiffusionDB \cite{Wang2022DiffusionDBAL} and set the threshold of UFID to $0.776$. For DisDet, we compute the the distribution discrepancy of the noise input of 3,000 benign samples from DiffusionDB and set the threshold of DisDet to 0.00176.

\textbf{Metrics.} We compute the F1 Score and Area Under the ROC Curve (AUC) for each detection method across all backdoor attack scenarios. The threshold of calculating F1 Score is selected based on train set, while AUC reflects the feature extraction ability that avoids the impact of threshold selection. Additionally, we report the number of generations required per sample and the detection inference time needed by each method.

\begin{table*}[]
\centering
\caption{Comparison of the proposed method with current methods on F1 score (\%) and AUC (\%). The top two results on each metric are \textbf{bolded} and \underline{underlined}, respectively.}
\label{tab:detection results}
\scalebox{0.84}{
\begin{tabular}{c|cccccccccccc|cc}
\toprule
                                  & \multicolumn{6}{c}{\textbf{F1 Score (\%) ↑}}                                                        & \multicolumn{6}{c|}{\textbf{AUC (\%) ↑}}                                                             &                                                                                    &                                                                                     \\ \cmidrule(lr){2-7} \cmidrule(lr){8-13}
\multirow{-2.5}{*}{\textbf{Method}} & Rickrolling    & Villan         & EvilEdit       & IBA            & BadT2I         & PersonBA    & Rickrolling    & Villan         & EvilEdit       & IBA            & BadT2I         & PersonBA    & \multirow{-2.5}{*}{\textbf{\begin{tabular}[|c]{@{}c@{}}Avg\\ F1 (\%) ↑\end{tabular}}} & \multirow{-2.5}{*}{\textbf{\begin{tabular}[c]{@{}c@{}}Avg\\ AUC (\%) ↑\end{tabular}}} \\ \hline
 DisDet \cite{sui2025disdet}                           & 0.0            & 88.57          & 0.0           & 0.0       & 5.83           & 0.0          & 51.57          & 93.51         & 54.52          & 50.42          & 53.35          & 70.30          & 15.73                                                                         & 62.28                                                                            \\
FTT \cite{wang2024t2ishield}
                              & 96.94          & 80.00          & 7.75           & 0.71           & 22.22          & 19.01          & \textbf{99.98} & 88.91          & 57.06          & 94.12          & 59.86          & 55.71          & 37.77                                                                              & 75.94                                                                               \\
CDA \cite{wang2024t2ishield}
                              & {\ul 97.60}    & 88.02          & 1.71           & 0.0            & 0.0            & 2.14           & 99.57          & 95.80          & 64.45          & 81.46          & 50.27          & 70.57          & 31.58                                                                              & 77.02                                                                               \\
UFID \cite{guan2024ufid}                             & 81.78          & 65.71          & 19.67          & 0.0            & 5.19           & 29.36          & 98.32          & {\ul 99.36}    & 53.13          & 65.59          & \textbf{70.12} & 63.21          & 33.62                                                                              & 74.96                                                                               \\
\rowcolor[HTML]{F2F2F2} 
DAA-I                             & 96.47          & {\ul 95.16}    & \textbf{75.89} & {\ul 54.98}    & {\ul 46.08}    & \textbf{72.00} & 98.14          & 99.20          & {\ul 78.07}    & {\ul 95.11}    & 60.35          & {\ul 73.95}    & {\ul 73.43}                                                                        & {\ul 84.14}                                                                         \\
\rowcolor[HTML]{F2F2F2} 
DAA-S                             & \textbf{97.44} & \textbf{98.95} & {\ul 73.68}    & \textbf{72.03} & \textbf{62.82} & {\ul 70.72}    & {\ul 99.80}    & \textbf{99.49} & \textbf{80.19} & \textbf{97.54} & {\ul 61.26}    & \textbf{79.35} & \textbf{79.27}                                                                     & \textbf{86.27}                                                                      \\ \bottomrule
\end{tabular}
}
\end{table*}

\begin{figure*}[tb]
  \centering
  \subfloat[DAA-I.]{\includegraphics[width = 0.48\textwidth]{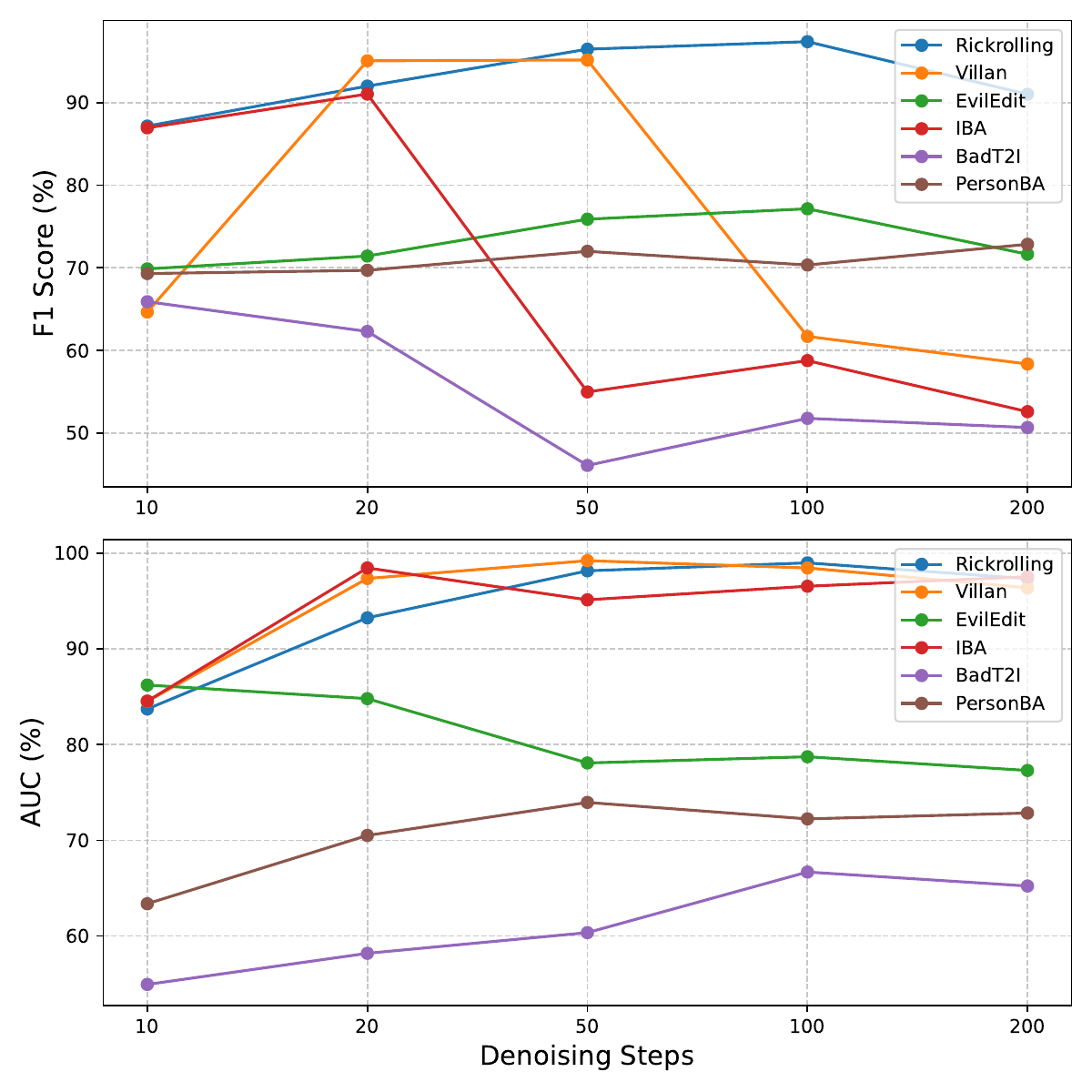}\label{fig:Sensitivity_to_steps_IND}}
  \hfill
  \subfloat[DAA-S.]{\includegraphics[width = 0.48\textwidth]{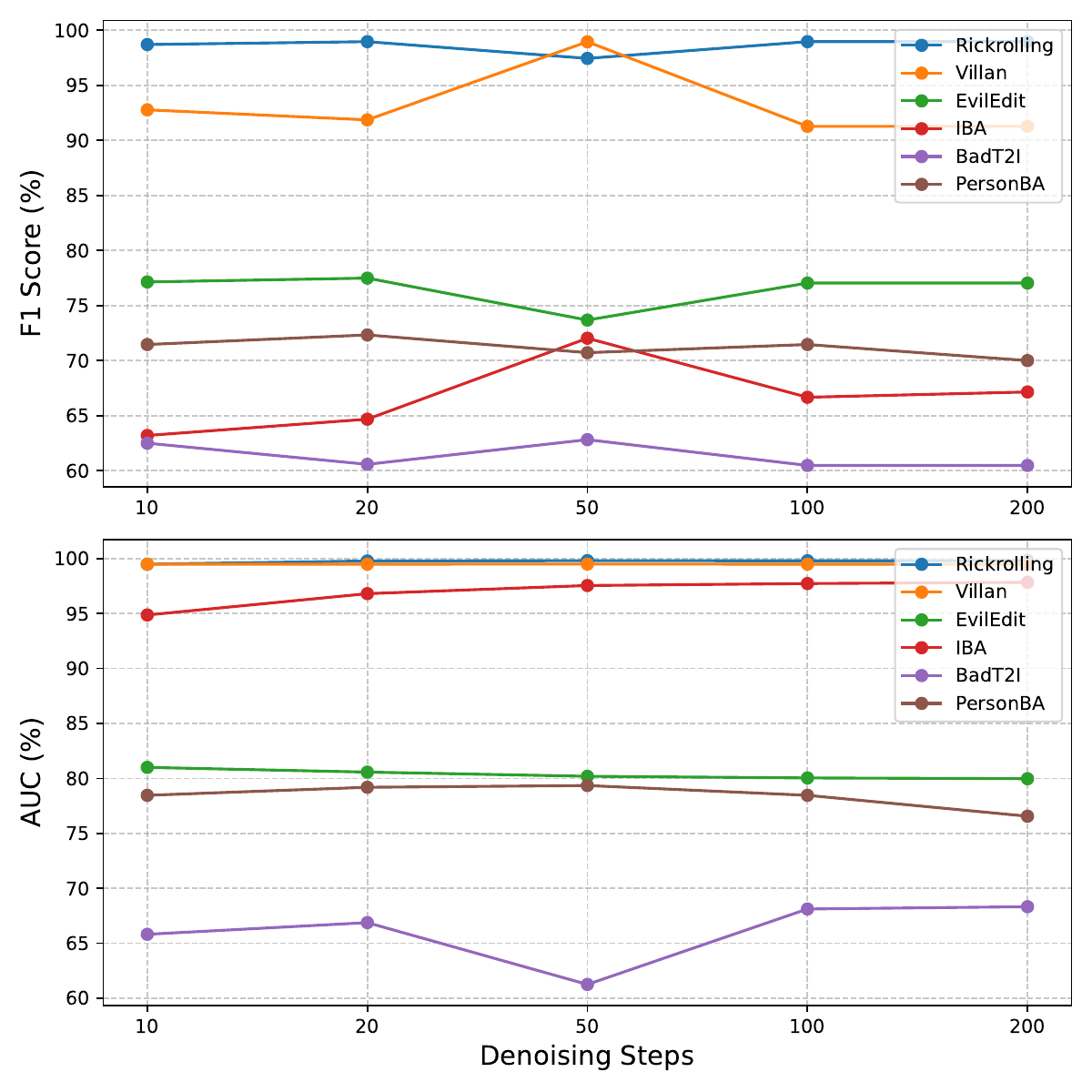}\label{fig:Sensitivity_to_steps_System}}
  \caption{Sensitivity to denoising steps. \textbf{\textit{(a)}} The line plots of DAA-I. \textbf{\textit{(b)}} The line plots of DAA-S. The upper plots the F1 score (\%) corresponding to the different denoising steps, and the lower plots represent the AUC (\%) corresponding to the different denoising steps. }
  \label{fig:Sensitivity_to_steps}
\end{figure*}

\textbf{Implementation details.} We train both $\text{DAA-I}$ and $\text{DAA-S}$ on train set to find the best $t$, $n$, $\alpha_I$ and $\alpha_S$. For the $\text{DAA-I}$, we set $t=3$ and $s=2$. For the $\text{DAA-S}$, we set $t=1$ and $s=5$. The thresholds $\alpha_I$ and $\alpha_S$ are set to $-0.0011$ and $-0.0053$, respectively. Besides, the coupling strength $c$ is set to 1. The dynamical matrix $F=\{-1,-1,\dots,-10\}$.  Denoising time steps $T$ for each sample is set to 50. The size of an attention map $D$ is $16\times16$.

\begin{figure*}[tb]
  \centering
  \subfloat[DAA-I.]{\includegraphics[width = 0.48\textwidth]{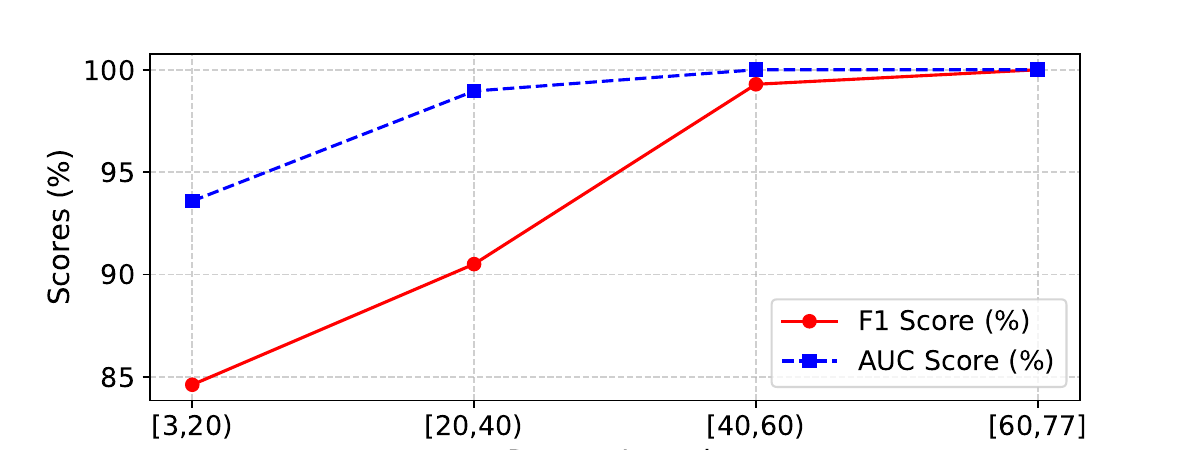}\label{fig:Sensitivity_to_length_IND}}
  \hfill
  \subfloat[DAA-S.]{\includegraphics[width = 0.49\textwidth]{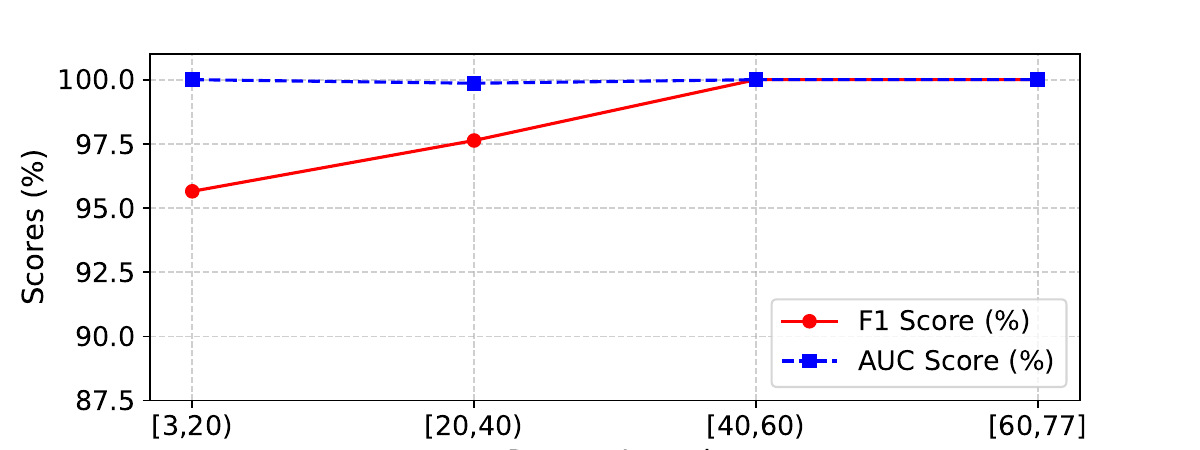}\label{fig:Sensitivity_to_length_System}}
  \caption{Sensitivity to prompt length of \textbf{\textit{(a)}} DAA-I and \textbf{\textit{(b)}} DAA-S. The red lines represent F1 Score (\%), and the blue lines represent AUC (\%), plotted against different prompt length spans: [3,20), [20,40), [40,60), and [60,77]. }
  \label{fig:Sensitivity_to_length}
\end{figure*}

\begin{figure*}[tb]
  \centering
  \subfloat[F1 Score.]{\includegraphics[width = 0.48\textwidth]{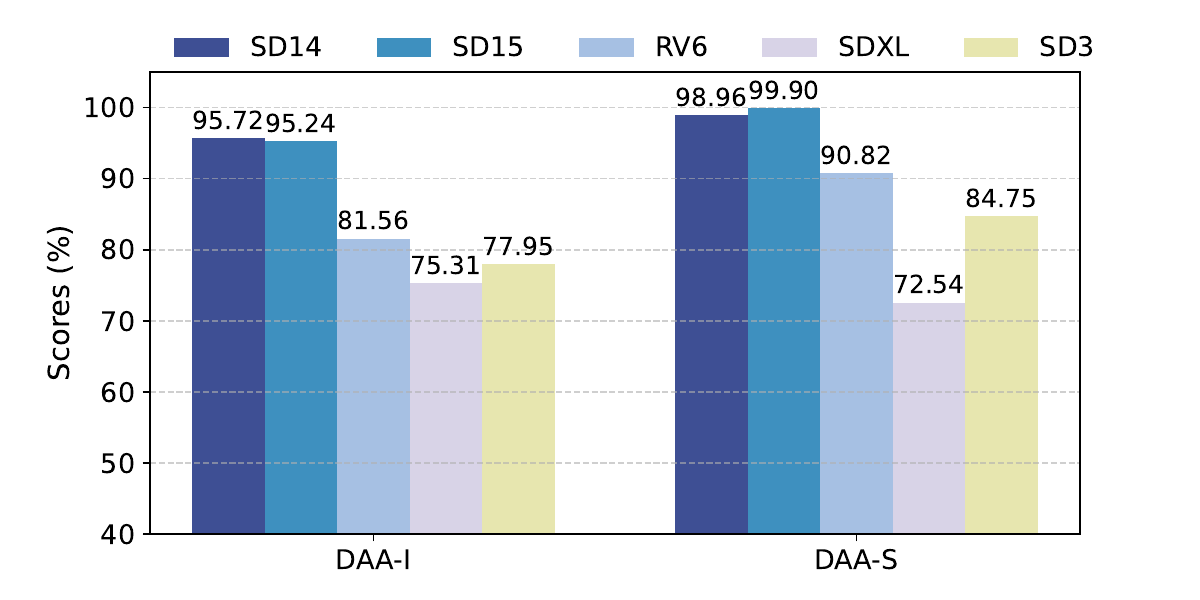}\label{fig:Sensitivity_to_base_model_IND}}
  \hfill
  \subfloat[AUC.]{\includegraphics[width = 0.48\textwidth]{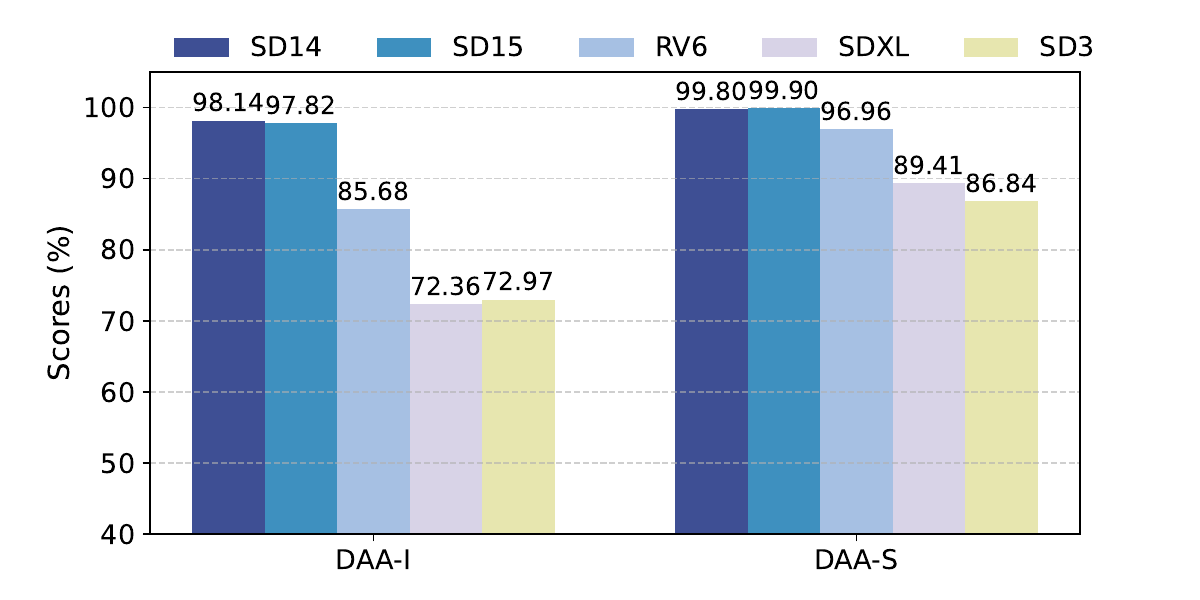}\label{fig:Sensitivity_to_base_model_System}}
  
  \caption{Sensitivity to the base model in terms of \textbf{\textit{(a)}} F1 Score (\%) and \textbf{\textit{(b)}} AUC (\%). The bars represent results on SD14 \cite{Rombach2021HighResolutionIS}, SD15 \cite{Rombach2021HighResolutionIS},  RV6 \cite{RealisticVision}, SDXL \cite{podell2024sdxl} and SD3 \cite{SD3}.}
  \label{fig:Sensitivity_to_base_model}
\end{figure*}

\subsection{Detection Results} \label{detection results}

Table \ref{tab:detection results} shows the comparison of our proposed $\text{DAA-I}$ and $\text{DAA-S}$ with state-of-the-art (SOTA) methods in terms of F1 Score and AUC. First, we observe that baseline methods \cite{wang2024t2ishield,guan2024ufid} are effective in detecting classic backdoor attack methods, \textit{i.e.}, Rickrolling \cite{Struppek2022RickrollingTA} and Villan \cite{Chou2023VillanDiffusionAU}. However, their performance significantly degrades on the latest backdoor attacks, including EvilEdit \cite{wang2024eviledit}, IBA \cite{zhang2025IBA}, BadT2I \cite{10.1145/3581783.3612108} and PersonBA \cite{Huang2023PersonalizationAA}. This decline is primarily due to the fact the static anomalies on which these detections rely are effectively mitigated by these attack methods. Besides, while DisDet \cite{sui2025disdet} achieve nearly 100\% accuary in detecting noise-like backdoor, it suffers a significant performance drop when facing text-based triggers in text-to-image backdoor attacks, achieving only average F1 Score of 15.73\% and average AUC of 62.28\%. This is because the textual trigger does not significantly alter the noise distribution, causing this detection method to fail in detecting backdoor samples in text-to-image diffusion models. In contrast, our DAA yields competitive results in detecting classical backdoors and exhibits superior performance in newer attack scenarios. DAA-I achieves an average F1 Score of 73.43\% and an average AUC of 84.14\%. DAA-S achieves the best detection performance with an average F1 Score of 79.27\% and an average AUC of 86.27\%. Both methods significantly outperforming the previous state-of-the-art F1 Score of 37.77\% and AUC of 77.02\%, demonstrating the effectiveness by leveraging dynamic attention features for backdoor detection. Furthermore, the improvement observed with DAA-S underscores the role of spatial correlations among attention maps in enhancing detection performance.

\subsection{Sensitivity to Generation Parameters} \label{sensitivity}
Since the attention features of a backdoor sample may be influenced by sample length, denoising steps, and the type of base model, we investigate the sensitivity of DAA to these generation parameters in this section.

\textbf{Sensitivity to denoising steps.} 
Specifically, for each attack scenario, we perform denoising for 10, 20, 50, 100, and 200 steps, respectively. We then calculate the detection performance across the backdoor scenarios for each denoising step.  Fig. \ref{fig:Sensitivity_to_steps} shows the line plots of DAA's performance across various denoising steps in terms of F1 and AUC. 
The upper plots represent the sensitivity of the F1 score to the number of denoising steps, while the lower plots show the sensitivity of the AUC. We observe that DAA-I maintains consistent performance across samples at different denoising steps. In contrast, DAA-S demonstrates remarkable stability, with its performance curve remaining almost flat. These findings suggest that DAA-S is less sensitive to the number of denoising steps compared to DAA-I, highlighting its robustness across varying number of denoising.

\textbf{Sensitivity to prompt length.}
Since different prompt lengths lead to varying scales of attention maps, we analyze the sensitivity of DAA to the prompt length. In this experiment, we use Rickrolling as the backdoor attack method and divide the prompt lengths into four spans: [3, 20), [20, 40), [40, 60), and  [60, 77]. The detection results on each span are shown in Fig. \ref{fig:Sensitivity_to_length}. As observed, DAA-S outperforms DAA-I across all prompt length spans, with both the F1 Score and AUC of DAA-S exceeding 95\%. Besides, we surprisingly find that detection performance improves as the prompt length increases. When the prompt length exceeds 60, both of methods achieve a F1 Score and AUC of nearly 100\%. We speculate that longer prompts require the backdoor trigger to exert a stronger influence on other tokens, leading to more pronounced anomalous features that are easier to detect.

\textbf{Sensitivity to base model types.} Given the plug-and-play nature of the backdoor-infected text encoder across various base models \cite{Struppek2022RickrollingTA}, we investigate the performance of DAA in detecting backdoor samples generated by different base models. In this experiment, we employ a text encoder injected with the Rickrolling trigger \cite{Struppek2022RickrollingTA} and replace the original benign text encoders of SD14 \cite{Rombach2021HighResolutionIS}, SD15 \cite{Rombach2021HighResolutionIS}, and Realistic Vision V6 (RV6) \cite{RealisticVision}. We also test a larger-scale model, \textit{i.e.}, SDXL \cite{podell2024sdxl} as well as a DiT-based model, \textit{i.e.}, SD3 \cite{SD3}.
We use the parameters trained on SD14 and directly apply them to other backbone models. As shown in Fig. \ref{fig:Sensitivity_to_base_model}, DAA achieves more than 72\% in terms of F1 score in backdoor detection across all models. On SD15, the F1 results are consistent with those on SD14, achieving a F1 Score of 95.24\% for DAA-I and 99.90\% for DAA-S. We also observe some performance degradation on other models. This may be because the feature distributions of samples vary across different models, making the original thresholds less effective. Nevertheless, we find that DAA-S consistently maintained AUC values above 86\% across all models, indicating that it can still effectively capture the discriminative features between backdoor and benign samples. The results indicate that the $<$EOS$>$ token's anomalous behavior is a strong feature for detection, regardless of the specific base model. Furthermore, the superior performance of DAA-S suggests that it exhibits less sensitivity to base model types compared to DAA-I.

\begin{table}[t]
\centering
\caption{Efficiency analysis of detection methods. $U$ and $V$ represent the number of image generations and denoising time steps per sample, respectively. $T_{DI}$ represents the detection model inference time. $T_{All}$ is the whole time cost per detection.}
\label{tab:efficiency}
\scalebox{0.93}{
\begin{tabular}{ccccc}
\hline
\textbf{Method} & \textbf{$U\times V$ $\downarrow$} & \textbf{$T_{DI}$ (ms) $\downarrow$}  & \textbf{$T_{All}$ (s) $\downarrow$} & \textbf{Avg. AUC (\%) $\uparrow$} \\ \hline
 DisDet  \cite{sui2025disdet}          & 1$\times$1                                    & 0.27                    & 0.06    & 62.28            \\
FTT \cite{wang2024t2ishield}             & 1$\times$50                                       & 0.09              &     3.18      & 75.94                 \\
CDA  \cite{wang2024t2ishield}           & 1$\times$50                                       & 12.59               &     3.21      & 77.02               \\
UFID  \cite{guan2024ufid}          & 15$\times$50                                    & 101.70                    & 47.22          & 74.96       \\

\rowcolor[HTML]{F2F2F2} 
DAA-I           & 1$\times$5                                        & 0.27                            &     0.32     &  84.14  \\
\rowcolor[HTML]{F2F2F2} 
DAA-S           & 1$\times$6                                        & 39.06                           &      0.41     & 86.27  \\ \hline
\end{tabular}
}
\end{table}

\subsection{Efficiency} \label{efficiency}

In this section, we compare the efficiency of DAA with several baseline methods. We conduct the experiments on a RTX4090-24GB GPU and the detailed implementation conditions are provided in Appendix-A. We apply different detection methods on 100 samples and calculate the average time cost per sample. For a backdoor detection method on the text-to-image diffusion model, the all time cost $T_{All}$ per sample is proportional to:
\begin{equation}
    T_{All} \propto (U \times V \times T_{DE} + T_{DI}),
\end{equation}
where the $U$ denotes the number of images to be generated per sample, and $V$ refers to the number of denoising steps per image. $T_{DE}$ is the time cost per denoising step and $T_{DI}$ is the detection model inference time. Here, we set $V=50$. 

For T2IShield, it requires $U=1$ generation and additional detection inference time for the ``Assimilation Phenomenon.” For UFID, it takes $U=15$ generations and additional detection inference time to compute graph diversity. For Disdet, since it focuses on the distribution discrepancy of the denoising noise, it only requires $V=1$. Similarly, DAA only also focuses on the early stages in denoising process, the detection is completed before the image generation. 

We present the detailed time cost in Table \ref{tab:efficiency}. As can be seen, DisDet achieves the fastest speed on $T_{DI}$, only requiring 0.27ms for calculating the feature and taking 0.06s to complete the whole detection. However, the average AUC of Disdet only achieves 62.28\%, making it fail to detect backdoor samples. DAA-I takes 0.27 ms to model the dynamic feature of the attention maps, while DAA-S takes 39.06 ms. Although DAA requires more time cost on detection inference, since the $T_{DE} \gg T_{DI}$, DAA achieve a significant low time consumption while the best detection performance compared to previous methods. DAA-I requires about 10\% time cost of FTT and DAA-S requires about 13\%. These results highlight the efficiency advantage of DAA over existing methods, making it a practical solution for effective and efficient backdoor detection in T2I diffusion models.

\begin{figure*}[tb]
  \centering
  \subfloat[DAA-I.]{\includegraphics[width = 0.48\textwidth]{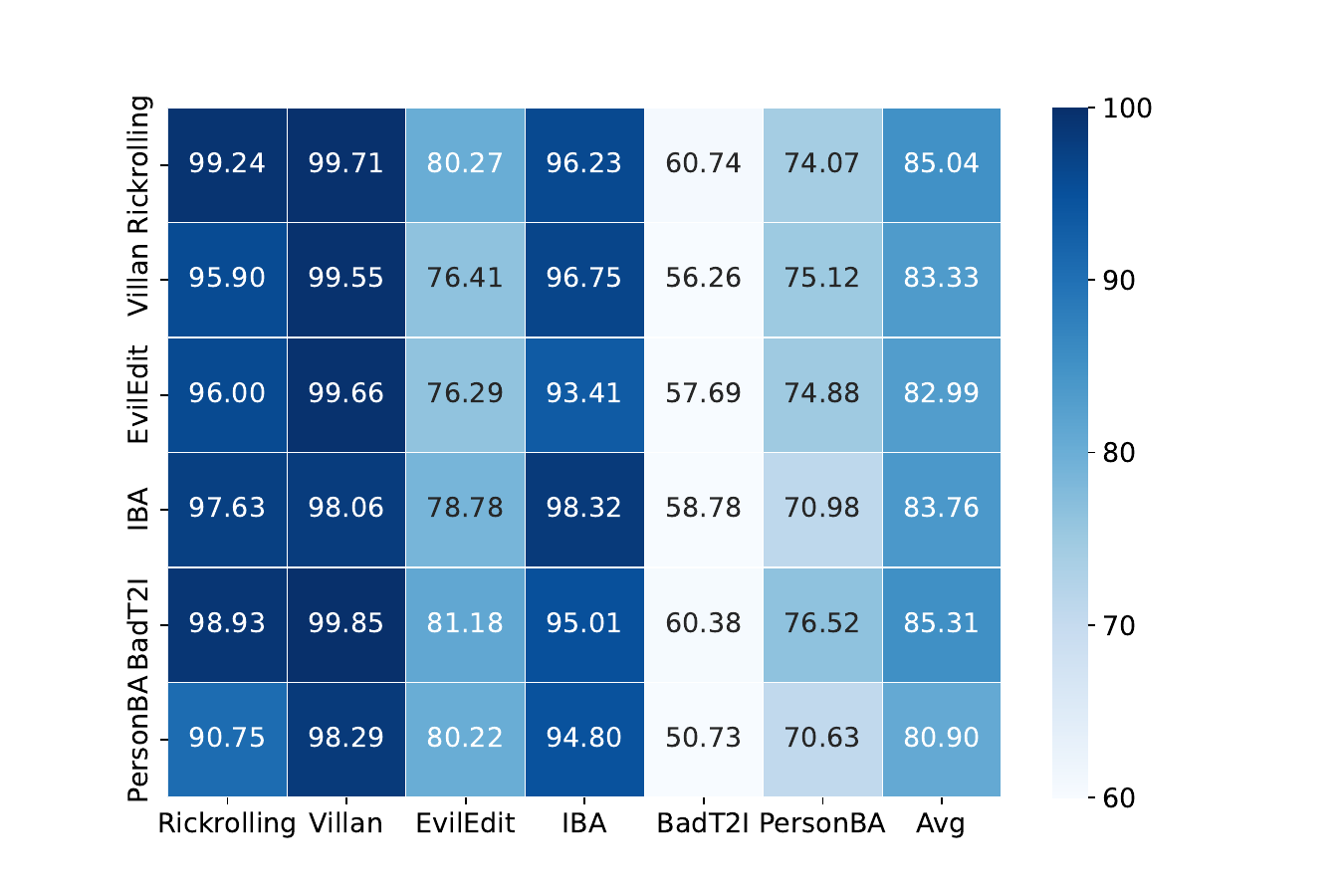}\label{fig:DAA-IND_generalization}}
  \hfill
  \subfloat[DAA-S.]{\includegraphics[width = 0.48\textwidth]{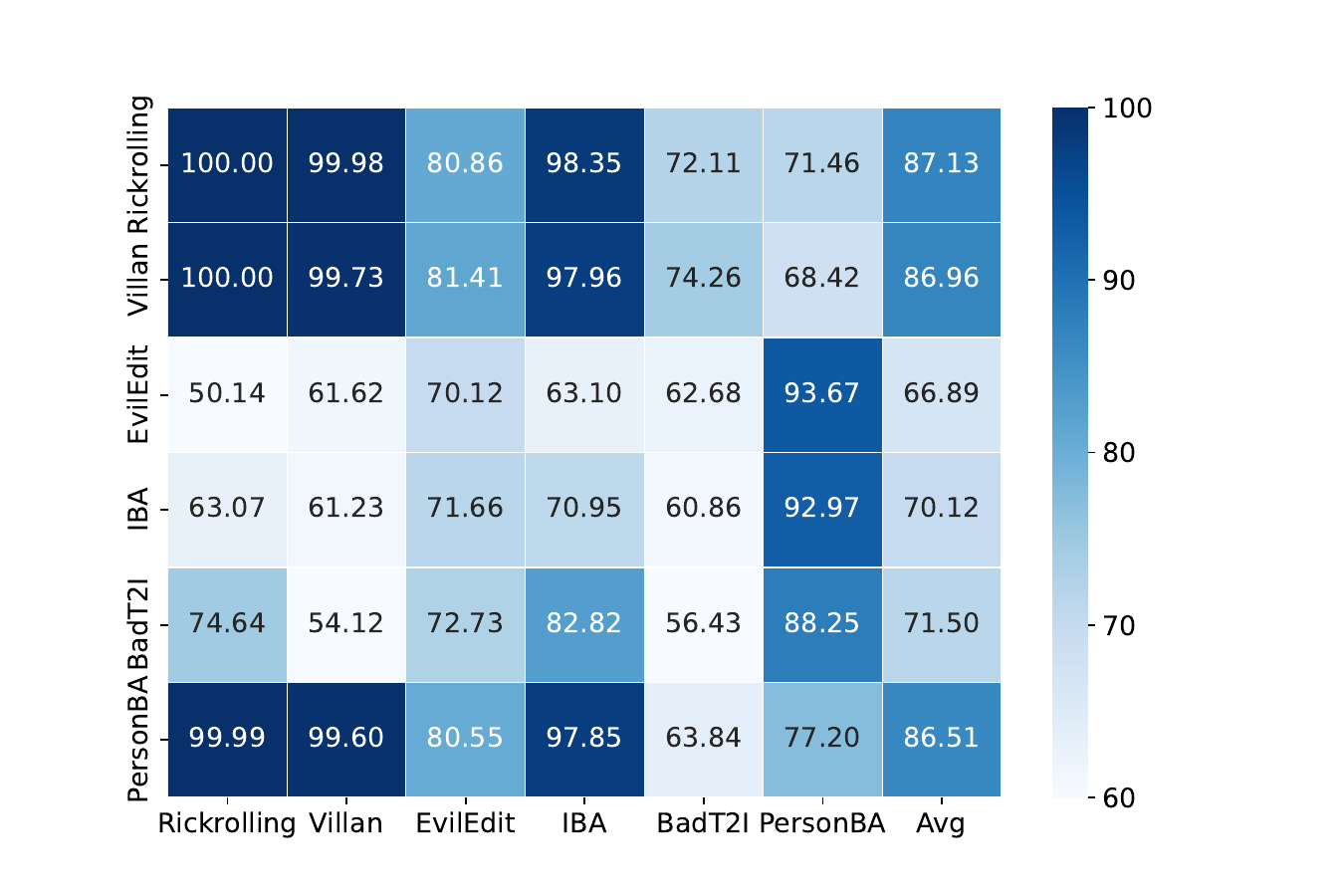}\label{fig:DAA-System_generalization}}
  \caption{The generalization of DAA. Each number in the rows represents the score of AUC (\%) for the corresponding attack scenario, with the last column showing the average AUC across six scenarios.}
  \label{fig:generalization}
\end{figure*}

\begin{table*}[]
\centering
\caption{Ablation study on start time step $t$ and time span $s$. Each value represents the F1 Score (\%) of each parameter combination. The best results are \textbf{bolded}.}
\label{tab:tsti}
\scalebox{1.3}{
\begin{tabular}{c|c|ccccccccc}
\toprule
\textbf{Method}         & \textbf{Start Time} & \textbf{s=1} & \textbf{s=2} & \textbf{s=3} & \textbf{s=4} & \textbf{s=5} & \textbf{s=6} & \textbf{s=7} & \textbf{s=8} & \textbf{s=9} \\ \midrule
                        & t=0 & 69.72 & 71.49 & 73.12 & 75.78 & 76.50 & 76.52 & 76.37 & 76.13 & 76.09 \\
                        & t=1 & 66.66 & 74.47 & 76.19 & 76.12 & 76.05 & 76.38 & 75.95 & 75.39 & 75.13 \\
                        & t=2 & 74.53 & 76.24 & 75.79 & 76.10 & 76.33 & 75.75 & 75.84 & 75.83 & 75.11 \\
\multirow{-4}{*}{DAA-I} & t=3 & 73.28 & \textbf{76.55} & 76.37 & 75.97 & 75.57 & 75.70 & 74.84 & 74.88 & 73.70 \\ \midrule
                        & t=0 & 71.87 & 76.95 & 75.47 & 76.57 & 76.81 & 77.45 & 77.87 & 77.96 & 78.37 \\
                        & t=1 & 77.26 & 76.83 & 77.24 & 78.34 & \textbf{79.72} & 78.86 & 79.32 & 79.63 & 79.71 \\
                        & t=2 & 77.80 & 78.74 & 78.59 & 78.00 & 4.67  & 6.26  & 7.39  & 8.35  & 7.93  \\
\multirow{-4}{*}{DAA-S} & t=3 & 78.00 & 4.43  & 4.47  & 5.23  & 8.38  & 8.15  & 8.49  & 8.09  & 10.39 \\ \bottomrule

\end{tabular}
}
\end{table*}

\subsection{Generalization} \label{generalization}
To study the generalization of the proposed methods, we train the models on one backdoor scenario and test them on others. Fig. \ref{fig:generalization} shows the generalization of DAA-I and DAA-S respectively. Each number in the rows represents the AUC value for the corresponding attack scenario, with the last column showing the average AUC across the six scenarios. Darker colors indicate higher values. As observed, the two methods exhibit different generalization capabilities. For DAA-I, the differences across each column are not significant, suggesting that the detection performance of this method is less sensitive to the distribution of the training data. We also notice that the detection performance on BadT2I is poor across all rows, which suggests that BadT2I is a challenging backdoor attack scenario for DAA-I.

Additionally, we observe that DAA-S achieves the best generalization performance when trained on the Rickrolling \cite{Struppek2022RickrollingTA}, Villan \cite{Chou2023VillanDiffusionAU} and PersonBA \cite{Huang2023PersonalizationAA} attack scenarios, with average AUC scores of 87.13\%, 86.96\% and 86.51\%, respectively. However, in the remaining three scenarios, DAA-S exhibits poorer generalization performance compared to DAA-I. This highlights the variability in DAA-S's performance across different attack types, suggesting that while it excels in certain contexts, its generalization capability is not as robust as that of DAA-I.

\begin{table*}[t]
\centering
\caption{Ablation study on $\gamma_L$. The top two results on each metric are \textbf{bolded} and \underline{underlined}, respectively.}
\label{tab:gamma_effect}
\scalebox{0.84}{
\begin{tabular}{c|cccccc|cccccc|cc}
\toprule
\multirow{2.5}{*}{\textbf{DAA-S}} 
& \multicolumn{6}{c|}{\textbf{F1 Score (\%) $\uparrow$}} 
& \multicolumn{6}{c|}{\textbf{AUC (\%) $\uparrow$}} 
& \multirow{2.5}{*}{\textbf{\begin{tabular}[c]{@{}c@{}}Avg\\ F1 (\%) $\uparrow$\end{tabular}}} 
& \multirow{2.5}{*}{\textbf{\begin{tabular}[c]{@{}c@{}}Avg\\ AUC (\%) $\uparrow$\end{tabular}}} 
\\ \cmidrule(lr){2-7} \cmidrule(lr){8-13}
& Rickrolling & Villan & EvilEdit & IBA & BadT2I & PersonBA 
& Rickrolling & Villan & EvilEdit & IBA & BadT2I & PersonBA 
& & \\ \midrule
$\gamma_L=-0.5$ & 96.94 & 93.66 & 74.10 & 66.67 & 60.22 & 71.34 & 97.96 & 99.38 & 82.45 & 96.29 & 58.54 & 87.88 & 77.16 & \underline{87.08} \\
$\gamma_L=-1$   & 96.94 & 95.68 & 73.68 & 67.15 & 61.54 & 74.42 & 98.17 & 99.48 & 82.15 & 96.60 & 57.95 & 87.32 & 78.24 & 86.95 \\
$\gamma_L=-2$   & 97.19 & 95.68 & 73.68 & 69.04 & 64.13 & 73.21 & 97.47 & 99.44 & 81.48 & 96.45 & 56.06 & 88.03 & 78.82 & 86.49 \\
$\gamma_L=-5$   & 97.44 & 89.71 & 71.92 & 76.32 & 64.13 & 70.54 & 97.23 & 98.98 & 82.93 & 95.52 & 58.58 & 89.29 & 78.18 & \textbf{87.09} \\
$\gamma_L=-10$  & 97.44 & 98.95 & 73.68 & 72.03 & 62.82 & 70.72 & 99.80 & 99.49 & 80.19 & 97.54 & 61.26 & 79.35 & \underline{79.27} & 86.27 \\
$\gamma_L=-15$  & 96.20 & 97.10 & 69.91 & 88.40 & 65.32 & 70.11 & 97.06 & 99.14 & 78.26 & 85.74 & 53.98 & 85.77 & \textbf{81.17} & 83.33 \\

\bottomrule
\end{tabular}
}
\end{table*}

\begin{table*}[]
\centering
\caption{Ablation study on $c$. The top two results on each metric are \textbf{bolded} and \underline{underlined}, respectively.}
\label{tab:c}
\scalebox{0.85}{
\begin{tabular}{c|cccccc|cccccc|cc}
\toprule
\multirow{2.5}{*}{\textbf{DAA-S}} & \multicolumn{6}{c|}{\textbf{F1 Score (\%) $\uparrow$}}                                      & \multicolumn{6}{c|}{\textbf{AUC (\%) $\uparrow$}}                                           & \multirow{2.5}{*}{\textbf{\begin{tabular}[c]{@{}c@{}}Avg\\ F1 (\%) $\uparrow$\end{tabular}}} & \multirow{2.5}{*}{\textbf{\begin{tabular}[c]{@{}c@{}}Avg\\ AUC (\%) $\uparrow$\end{tabular}}} \\ \cmidrule(lr){2-7} \cmidrule(lr){8-13}
                                & Rickrolling    & Villan         & EvilEdit       & IBA            & BadT2I         & PersonBA       & Rickrolling    & Villan         & EvilEdit       & IBA            & BadT2I         & PersonBA       &                                                                                 &                                                                                  \\ \midrule
$c=0.5$   & 95.24 & 97.63 & 70.81 & 78.37 & 63.92 & 72.69 & 97.22 & 99.43 & 81.27 & 95.75 & 57.71 & 82.03 & 79.78 & 85.40 \\
$c=1$     & 97.44 & 98.95 & 73.68 & 72.03 & 62.82 & 70.72 & 99.80 & 99.49 & 80.19 & 97.54 & 61.26 & 79.35 & 79.27 & \textbf{86.27} \\
$c=5$     & 96.18 & 96.08 & 71.44 & 82.69 & 62.92 & 71.69 & 98.76 & 99.42 & 79.84 & 97.80 & 55.42 & 80.56 & \underline{80.17} & 85.13 \\
$c=10$    & 97.19 & 96.79 & 73.38 & 78.18 & 63.69 & 71.98 & 98.31 & 99.46 & 81.12 & 97.78 & 55.94 & 82.49 & \textbf{80.20} & \underline{85.68} \\ \bottomrule
\end{tabular}
}
\end{table*}

\begin{table}[]
\centering
\caption{Ablation study on token selections.}
\label{tab:token-selection}
\scalebox{1.3}{
\begin{tabular}{cccc}
\hline
\multirow{2}{*}{\textbf{}} & \multicolumn{3}{c}{\textbf{Avg AUC (\%) $\uparrow$}}                         \\ \cline{2-4} 
                           & \textless{}EOS\textgreater{} & \textless{}BOS\textgreater{} & All Tokens  \\ \hline
\textbf{DAA-I}             & {\ul 84.14}                 & 67.48                        & 63.48 \\
\textbf{DAA-S}             & \textbf{86.27}               & 64.79                        & 70.15 \\ \hline
\end{tabular}
}
\end{table}

\begin{table*}[]
\centering
\caption{Ablation study on similarity metric. The top two results on each metric are \textbf{bolded} and \underline{underlined}, respectively.}
\label{tab:similarity metric}
\scalebox{0.86}{
\begin{tabular}{c|cccccc|cccccc|cc}
\toprule
\multirow{2.5}{*}{\textbf{DAA-S}} & \multicolumn{6}{c|}{\textbf{F1 Score (\%) $\uparrow$}}                                      & \multicolumn{6}{c|}{\textbf{AUC (\%) $\uparrow$}}                                           & \multirow{2.5}{*}{\textbf{\begin{tabular}[c]{@{}c@{}}Avg\\ F1 (\%) $\uparrow$\end{tabular}}} & \multirow{2.5}{*}{\textbf{\begin{tabular}[c]{@{}c@{}}Avg\\ AUC (\%) $\uparrow$\end{tabular}}} \\ 
\cmidrule(lr){2-7} \cmidrule(lr){8-13}
& Rickrolling & Villan & EvilEdit & IBA & BadT2I & PersonBA & Rickrolling & Villan & EvilEdit & IBA & BadT2I & PersonBA & & \\ 
\midrule
SSIM  & 85.45 & 98.95 & 64.84 & 26.76 & 26.75 & 70.59 & 90.72 & 99.83 & 82.05 & 84.31 & 70.01 & 74.54 & 62.22 & \underline{83.58} \\
1-norm & 92.01 & 96.60 & 70.34 & 60.08 & 65.67 & 70.93 & 96.34 & 99.36 & 80.99 & 72.85 & 54.97 & 87.53 & \underline{75.94} & 82.01 \\
F-norm & 97.44 & 98.95 & 73.68 & 72.03 & 62.82 & 70.72 & 99.80 & 99.49 & 80.19 & 97.54 & 61.26 & 79.35 & \textbf{79.27} & \textbf{86.27} \\
\bottomrule
\end{tabular}
}
\end{table*}

\subsection{Ablation Study} \label{ablation study}
In this section, we conduct experiments to analyze the effect of the hyperparameters on our methods.

\textbf{Effect of start time step $t$ and time span $s$.} Recall that in Eq. \eqref{DAA-I} and Eq. \eqref{DAA-S}, $t$ and $s$ represent the start time step and time span, respectively. To evaluate their impact on DAA, we compute the F1 Score on the training set under different combinations of these parameters. As shown in Table \ref{tab:tsti}, DAA-I achieves the best F1 Score when $t=3$ and $s=2$. DAA-S achieves the best F1 Score when $t=1$ and $s=5$. We observe that the anomalous dynamic feature of the backdoor attack is more effectively captured during the earlier stage of the diffusion process. We believe this is due to the relatively intense attention evolution during the early diffusion process. As shown in Fig. \ref{fig:lvy}, the Lyapunov function derivative exhibits significant oscillations during the early 10 steps, implying that the evolution are more pronounced. The oscillation in early stages leads to more distinct dynamic features between benign and backdoor samples in the early stages.


\textbf{Effect of $\gamma_L$.} Recall that $\gamma_i, i\in[1,L]$ are the stability parameters for each node in the DAA-S dynamical system, with larger absolute values indicating greater stability. We perform an ablation experiment on $\gamma_L$ to evaluate its impact on detection performance. As shown in Table \ref{tab:gamma_effect}, We observe that DAA-S exhibits similar F1 Scores and AUCs under different values of $\gamma_L$, demonstrating that the algorithm is insensitive to this parameter. Here, we select $\gamma_L=-10$ as the optimal parameter to achieve the best balance between the F1 Score and AUC. 

\textbf{Effect of $c$.} Recall that $c$ serves as the coupling coefficient in the dynamical system, balancing the influence between correlation features and node features within the system. Table \ref{tab:c} presents the results of the ablation experiment. As shown, DAA-S achieves the best F1 Score of 80.17\% when $c=5$, while achieving the best AUC of 86.27\% when $c=1$. Finally, we select $c=1$ as the optimal parameter to achieve the best AUC performance.

\textbf{Effect of token selections.} Here, we explore the impact of different token in calculating the Relative Evolve Rate (RER) in Eq. \eqref{rer}. Specifically, we consider two conditions: 1) using the $<$BOS$>$ token instead of the $<$EOS$>$ token, and 2) using the average Evolve Rate (ER) feature of all tokens. The average AUC for the six backdoor attack scenarios is shown in Table \ref{tab:token-selection}. As observed, the AUC for both conditions are around 65\%, indicating that these features based on these tokens are not as effective as the features from $<$EOS$>$ token for backdoor detection. This result highlights the unique role of the $<$EOS$>$ token in detecting backdoor samples.

\textbf{Effect of similarity metric.} In Eq. \eqref{similarity}, we use the F-norm to represent the similarity between two attention maps. Here, we examine the impact of using 1-norm and SSIM \cite{ssim} as similarity metrics on detection performance. As shown in Table \ref{tab:similarity metric}, the F-norm achieves the best performance, with an F1 score of 79.27\% and an AUC of 86.27\%. Therefore, we select the F-norm as the optimal similarity metric.

\begin{table}[]
\centering
\caption{Backdoor detection results against the EvilEdit and its adaptive attack version.}
\label{tab:adaptive attack}
\scalebox{1.1}{
\begin{tabular}{ccccc}
\toprule
\multirow{2}{*}{\textbf{}}                                             & \multirow{2.5}{*}{\textbf{FID $\downarrow$}} & \multirow{2.5}{*}{\textbf{ASR (\%) $\uparrow$}} & \multicolumn{2}{c}{\textbf{AUC (\%) $\uparrow$}} \\ \cmidrule(lr){4-5}
&            &                  & DAA-I                                      &  DAA-S                               \\ \midrule
Origin SD                                                              & 19.07                & -                        & -                                    & -                           \\ \hline
EvilEdit                                                               &  18.82           &  80.00          &  78.55                                  & 85.95                           \\ 
\begin{tabular}[c]{@{}c@{}}Adaptive Attack\end{tabular} & 18.58             &  56.00           &   74.25                                 &   82.83                              \\ \bottomrule
\end{tabular}
}
\end{table}

\subsection{Defense Robustness against Potential Adaptive Attack} \label{defense robustness}

In this section, we test the robustness of our methods against potential adaptive attacks. We assume that the attacker is aware of the backdoor detection mechanism of DAA, which leverages the anomalous attention of the $<$EOS$>$ token. The adaptive attack aims to avoid altering the attention of the $<$EOS$>$ token during backdoor injection. To achieve this, we propose a method based on EvilEdit \cite{wang2024eviledit}. Recall that EvilEdit injects backdoors into cross-attention via a closed-form solution:
\begin{equation}
    W^*=(Wc^{ta}{c^{tr}}^T+\lambda W)(c^{tr}{c^{tr}}^T+\lambda \mathbb{I})^{-1},
    \label{eviledit_form}
\end{equation}
where $c^{tr}$ and $c^{ta}$ are the trigger embedding and backdoor target by CLIP text encoder \cite{Radford2021LearningTV}, respectively. $W$ is the original cross-attention parameter, while $W^*$ is the edited parameter. $\lambda=1$ is the regularization hyper-parameter. $\mathbb{I}$ is the identity matrix. We perform the adaptive attack by removing the corresponding embedding of the $<$EOS$>$ token:
\begin{equation}
   c^{tr}=[c^{tr}_1,c^{tr}_2,\dots, c^{tr}_L,\dots,c^{tr}_N], 
\end{equation}
\begin{equation}
c^{tr}_{new} = [c^{tr}_1,c^{tr}_2,\dots, c^{tr}_{L-1},c^{tr}_{L+1},\dots,c^{tr}_N],
\end{equation}
where $L$ is the position of the $<$EOS$>$ token, $N=77$ is the token length. The same operation is applied to $c^{ta}$:
\begin{equation}
c^{ta}_{new} = [c^{ta}_1,c^{ta}_2,\dots, c^{ta}_{L-1},c^{ta}_{L+1},\dots,c^{ta}_N].
\end{equation}

Then, the backdoored parameter $W^*_{new}$ with adaptive attack is computed by:
\begin{equation}
    W^*_{new}=(Wc^{ta}_{new}{c^{tr}_{new}}^T+\lambda W)(c^{tr}_{new}{c^{tr}_{new}}^T+\lambda \mathbb{I})^{-1}.
    \label{adaptive form}
\end{equation}

We compare the results of EvilEdit and its adaptive attack version in Table \ref{tab:adaptive attack}. First, to assess the performance of the backdoor model on benign samples, we compute the Frechet Inception Distance (FID) \cite{parmar2021cleanfid} using the COCO-30k validation subset \cite{Lin2014MicrosoftCC}. It can be observed that both models injected with a backdoor maintain the same generation performance as the original SD model. However, we find that the ASR of EvilEdit with such adaptive attack suffers a decline, dropping from 80.00\% to 56.00\%. The decline suggests the difficulty in balancing the attack capacity and the resistance to defense. To ensure a fair comparison, only the successfully attacked samples from both methods are considered in calculating the detection AUC. We observe that the detection performance of DAA slightly declines when facing adaptive attacks. The detection AUC of DAA-I decreases from 78.55\% to 74.25\%, while the detection AUC of DAA-S decreases from 85.95\% to 82.83\%.  The results demonstrate that DAA still maintains the ability in detecting potential adaptive attack samples.

\section{Conclusion}
This paper introduces DAA, a novel approach for detecting backdoor samples in text-to-image diffusion models by leveraging dynamic attention analysis. Specifically, we find the dynamic anomaly of the $<$EOS$>$ token in backdoor samples and propose two effective detection methods based on this anomaly. Experimental results on six representative backdoor attack scenarios demonstrate the efficiency and effectiveness of DAA. Our work offers a new perspective on dynamic attention analysis to understand the behavior of backdoor samples. We believe our work opens up the floor for future investigation in both attacks and defenses, and contributes to build safe and reliable diffusion models.

\bibliographystyle{IEEEtran}
\bibliography{main}

\begin{thebibliography}{10}
\providecommand{\url}[1]{#1}
\csname url@samestyle\endcsname
\providecommand{\newblock}{\relax}
\providecommand{\bibinfo}[2]{#2}
\providecommand{\BIBentrySTDinterwordspacing}{\spaceskip=0pt\relax}
\providecommand{\BIBentryALTinterwordstretchfactor}{4}
\providecommand{\BIBentryALTinterwordspacing}{\spaceskip=\fontdimen2\font plus
\BIBentryALTinterwordstretchfactor\fontdimen3\font minus \fontdimen4\font\relax}
\providecommand{\BIBforeignlanguage}[2]{{%
\expandafter\ifx\csname l@#1\endcsname\relax
\typeout{** WARNING: IEEEtran.bst: No hyphenation pattern has been}%
\typeout{** loaded for the language `#1'. Using the pattern for}%
\typeout{** the default language instead.}%
\else
\language=\csname l@#1\endcsname
\fi
#2}}
\providecommand{\BIBdecl}{\relax}
\BIBdecl

\bibitem{NEURIPS2020_4c5bcfec}
J.~Ho, A.~Jain, and P.~Abbeel, ``Denoising diffusion probabilistic models,'' in \emph{Advances in Neural Information Processing Systems (NeurIPS)}, vol.~33.\hskip 1em plus 0.5em minus 0.4em\relax Curran Associates, Inc., 2020, pp. 6840--6851.

\bibitem{song2021denoising}
J.~Song, C.~Meng, and S.~Ermon, ``Denoising diffusion implicit models,'' in \emph{International Conference on Learning Representations (ICLR)}, 2021.

\bibitem{NEURIPS2021_49ad23d1}
P.~Dhariwal and A.~Nichol, ``Diffusion models beat gans on image synthesis,'' in \emph{Advances in Neural Information Processing Systems (NeurIPS)}, vol.~34.\hskip 1em plus 0.5em minus 0.4em\relax Curran Associates, Inc., 2021, pp. 8780--8794.

\bibitem{ho2021classifierfree}
J.~Ho and T.~Salimans, ``Classifier-free diffusion guidance,'' in \emph{NeurIPS 2021 Workshop on Deep Generative Models and Downstream Applications}, 2021.

\bibitem{Ramesh2022HierarchicalTI}
A.~Ramesh, P.~Dhariwal, A.~Nichol, C.~Chu, and M.~Chen, ``Hierarchical text-conditional image generation with clip latents,'' \emph{arXiv preprint arXiv:2204.06125}, 2022.

\bibitem{Rombach2021HighResolutionIS}
R.~Rombach, A.~Blattmann, D.~Lorenz, P.~Esser, and B.~Ommer, ``High-resolution image synthesis with latent diffusion models,'' in \emph{Proceedings of the IEEE/CVF Conference on Computer Vision and Pattern Recognition (CVPR)}, 2021, pp. 10\,674--10\,685.

\bibitem{Yu2022ScalingAM}
J.~Yu, Y.~Xu, J.~Y. Koh, T.~Luong, G.~Baid, Z.~Wang, V.~Vasudevan, A.~Ku, Y.~Yang, B.~K. Ayan, B.~C. Hutchinson, W.~Han, Z.~Parekh, X.~Li, H.~Zhang, J.~Baldridge, and Y.~Wu, ``Scaling autoregressive models for content-rich text-to-image generation,'' \emph{Trans. Mach. Learn. Res.}, 2022.

\bibitem{esser2024sd3}
P.~Esser, S.~Kulal, A.~Blattmann, R.~Entezari, J.~M\"{u}ller, H.~Saini, Y.~Levi, D.~Lorenz, A.~Sauer, F.~Boesel, D.~Podell, T.~Dockhorn, Z.~English, and R.~Rombach, ``Scaling rectified flow transformers for high-resolution image synthesis,'' in \emph{International Conference on Machine Learning (ICML)}, ser. ICML'24.\hskip 1em plus 0.5em minus 0.4em\relax JMLR.org, 2024.

\bibitem{10081412}
F.-A. Croitoru, V.~Hondru, R.~T. Ionescu, and M.~Shah, ``Diffusion models in vision: A survey,'' \emph{IEEE Transactions on Pattern Analysis and Machine Intelligence (TPAMI)}, vol.~45, no.~9, pp. 10\,850--10\,869, 2023.

\bibitem{Kim2023StableVITONLS}
J.~Kim, G.~Gu, M.~Park, S.~Park, and J.~Choo, ``Stableviton: Learning semantic correspondence with latent diffusion model for virtual try-on,'' in \emph{Proceedings of the IEEE/CVF Conference on Computer Vision and Pattern Recognition (CVPR)}, 2024, pp. 8176--8185.

\bibitem{Zhu2023TryOnDiffusionAT}
L.~Zhu, D.~Yang, T.~L. Zhu, F.~A. Reda, W.~Chan, C.~Saharia, M.~Norouzi, and I.~Kemelmacher-Shlizerman, ``Tryondiffusion: A tale of two unets,'' in \emph{2023 IEEE/CVF Conference on Computer Vision and Pattern Recognition (CVPR)}, 2023, pp. 4606--4615.

\bibitem{Ghosh2022CanTB}
A.~Ghosh and G.~Fossas, ``Can there be art without an artist?'' \emph{arXiv preprint arXiv:2209.07667}, 2022.

\bibitem{10219843}
Z.~Wang, J.~Zhang, Z.~Ji, J.~Bai, and S.~Shan, ``Cclap: Controllable chinese landscape painting generation via latent diffusion model,'' in \emph{2023 IEEE International Conference on Multimedia and Expo (ICME)}, 2023, pp. 2117--2122.

\bibitem{KAZEROUNI2023102846}
A.~Kazerouni, E.~K. Aghdam, M.~Heidari, R.~Azad, M.~Fayyaz, I.~Hacihaliloglu, and D.~Merhof, ``Diffusion models in medical imaging: A comprehensive survey,'' \emph{Medical Image Analysis}, vol.~88, p. 102846, 2023.

\bibitem{Hertz2022PrompttoPromptIE}
A.~Hertz, R.~Mokady, J.~Tenenbaum, K.~Aberman, Y.~Pritch, and D.~Cohen-or, ``Prompt-to-prompt image editing with cross-attention control,'' in \emph{The Eleventh International Conference on Learning Representations (ICLR)}, 2023.

\bibitem{ruiz2023dreambooth}
N.~Ruiz, Y.~Li, V.~Jampani, Y.~Pritch, M.~Rubinstein, and K.~Aberman, ``Dreambooth: Fine tuning text-to-image diffusion models for subject-driven generation,'' in \emph{Proceedings of the IEEE/CVF Conference on Computer Vision and Pattern Recognition (CVPR)}, 2023.

\bibitem{Ma_He_Cun_Wang_Chen_Li_Chen_2024}
Y.~Ma, Y.~He, X.~Cun, X.~Wang, S.~Chen, X.~Li, and Q.~Chen, ``Follow your pose: Pose-guided text-to-video generation using pose-free videos,'' in \emph{Proceedings of the AAAI Conference on Artificial Intelligence (AAAI)}, vol.~38, no.~5, Mar. 2024, pp. 4117--4125.

\bibitem{Brooks_2023_CVPR}
T.~Brooks, A.~Holynski, and A.~A. Efros, ``Instructpix2pix: Learning to follow image editing instructions,'' in \emph{Proceedings of the IEEE/CVF Conference on Computer Vision and Pattern Recognition (CVPR)}, June 2023, pp. 18\,392--18\,402.

\bibitem{zhang2023adding}
L.~Zhang, A.~Rao, and M.~Agrawala, ``Adding conditional control to text-to-image diffusion models,'' in \emph{Proceedings of the IEEE/CVF International Conference on Computer Vision (ICCV)}, 2023, pp. 3836--3847.

\bibitem{zhang2025dysca}
J.~Zhang, Z.~Wang, M.~Lei, Z.~Yuan, B.~Yan, S.~Shan, and X.~Chen, ``Dysca: A dynamic and scalable benchmark for evaluating perception ability of {LVLM}s,'' in \emph{The Thirteenth International Conference on Learning Representations (ICLR)}, 2025.

\bibitem{Civitai}
``Civitai,'' \url{https://civitai.com}.

\bibitem{Midjourney}
``Midjourney,'' \url{www.midjourney.com}.

\bibitem{Struppek2022RickrollingTA}
L.~Struppek, D.~Hintersdorf, and K.~Kersting, ``Rickrolling the artist: Injecting backdoors into text encoders for text-to-image synthesis,'' in \emph{2023 IEEE/CVF International Conference on Computer Vision (ICCV)}, 2022, pp. 4561--4573.

\bibitem{Huang2023PersonalizationAA}
Y.~Huang, F.~Juefei-Xu, Q.~Guo, J.~Zhang, Y.~Wu, M.~Hu, T.~Li, G.~Pu, and Y.~Liu, ``Personalization as a shortcut for few-shot backdoor attack against text-to-image diffusion models,'' in \emph{Proceedings of the AAAI Conference on Artificial Intelligence (AAAI)}, Mar. 2024, pp. 21\,169--21\,178.

\bibitem{Chou2023VillanDiffusionAU}
S.-Y. Chou, P.-Y. Chen, and T.-Y. Ho, ``Villandiffusion: A unified backdoor attack framework for diffusion models,'' in \emph{Thirty-seventh Conference on Neural Information Processing Systems (NeurIPS)}, 2023.

\bibitem{Vice2023BAGMAB}
J.~Vice, N.~Akhtar, R.~Hartley, and A.~Mian, ``Bagm: A backdoor attack for manipulating text-to-image generative models,'' \emph{IEEE Transactions on Information Forensics and Security}, vol.~19, pp. 4865--4880, 2024.

\bibitem{Wu2023BackdooringTI}
Y.~Wu, J.~Zhang, F.~Kerschbaum, and T.~Zhang, ``Backdooring textual inversion for concept censorship,'' \emph{arXiv preprint arXiv:2308.10718}, 2023.

\bibitem{wang2024eviledit}
H.~Wang, S.~Guo, J.~He, K.~Chen, S.~Zhang, T.~Zhang, and T.~Xiang, ``Eviledit: Backdooring text-to-image diffusion models in one second,'' in \emph{ACM Multimedia (ACM MM)}, 2024.

\bibitem{Radford2021LearningTV}
A.~Radford, J.~W. Kim, C.~Hallacy, A.~Ramesh, G.~Goh, S.~Agarwal, G.~Sastry, A.~Askell, P.~Mishkin, J.~Clark, G.~Krueger, and I.~Sutskever, ``Learning transferable visual models from natural language supervision,'' in \emph{International Conference on Machine Learning (ICML)}, 2021.

\bibitem{Ronneberger2015UNetCN}
O.~Ronneberger, P.~Fischer, and T.~Brox, ``U-net: Convolutional networks for biomedical image segmentation,'' in \emph{Medical Image Computing and Computer-Assisted Intervention (MICCAI)}.\hskip 1em plus 0.5em minus 0.4em\relax Cham: Springer International Publishing, 2015, pp. 234--241.

\bibitem{Hu2021LoRALA}
J.~E. Hu, Y.~Shen, P.~Wallis, Z.~Allen-Zhu, Y.~Li, S.~Wang, and W.~Chen, ``Lora: Low-rank adaptation of large language models,'' \emph{arXiv preprint arXiv:2106.09685}, 2021.

\bibitem{10.1145/3581783.3612108}
S.~Zhai, Y.~Dong, Q.~Shen, S.~Pu, Y.~Fang, and H.~Su, ``Text-to-image diffusion models can be easily backdoored through multimodal data poisoning,'' in \emph{Proceedings of the 31st ACM International Conference on Multimedia (ACM MM)}, ser. MM '23.\hskip 1em plus 0.5em minus 0.4em\relax New York, NY, USA: Association for Computing Machinery, 2023, p. 1577–1587.

\bibitem{wang2024t2ishield}
Z.~Wang, J.~Zhang, S.~Shan, and X.~Chen, ``T2ishield: Defending against backdoors on text-to-image diffusion models,'' in \emph{Proceedings of the European Conference on Computer Vision (ECCV)}, 2024.

\bibitem{guan2024ufid}
Z.~Guan, M.~Hu, S.~Li, and A.~K. Vullikanti, ``Ufid: a unified framework for black-box input-level backdoor detection on diffusion models,'' in \emph{Proceedings of the Thirty-Ninth AAAI Conference on Artificial Intelligence (AAAI)}.\hskip 1em plus 0.5em minus 0.4em\relax AAAI Press, 2025.

\bibitem{zhang2025IBA}
J.~Zhang, Z.~Wang, S.~Shan, and X.~Chen, ``Towards invisible backdoor attack on text-to-image diffusion models,'' \emph{arXiv}, 2025.

\bibitem{pmlr-v37-sohl-dickstein15}
J.~Sohl-Dickstein, E.~Weiss, N.~Maheswaranathan, and S.~Ganguli, ``Deep unsupervised learning using nonequilibrium thermodynamics,'' in \emph{Proceedings of the 32nd International Conference on Machine Learning}, ser. Proceedings of Machine Learning Research, vol.~37.\hskip 1em plus 0.5em minus 0.4em\relax Lille, France: PMLR, 07--09 Jul 2015, pp. 2256--2265.

\bibitem{Vaswani2017AttentionIA}
A.~Vaswani, N.~M. Shazeer, N.~Parmar, J.~Uszkoreit, L.~Jones, A.~N. Gomez, L.~Kaiser, and I.~Polosukhin, ``Attention is all you need,'' in \emph{Neural Information Processing Systems (NeruIPS)}, 2017.

\bibitem{Lin2021CATCA}
H.~Lin, X.~Cheng, X.~Wu, F.~Yang, D.~Shen, Z.~Wang, Q.~Song, and W.~Yuan, ``Cat: Cross attention in vision transformer,'' \emph{2022 IEEE International Conference on Multimedia and Expo (ICME)}, pp. 1--6, 2021.

\bibitem{matrix_a}
R.~A.Horn and C.~R.Johnson, \emph{Matrix Analysis}.\hskip 1em plus 0.5em minus 0.4em\relax Cambridge University Press, 1985.

\bibitem{ZEMANOVA2006202}
L.~Zemanová, C.~Zhou, and J.~Kurths, ``Structural and functional clusters of complex brain networks,'' \emph{Physica D: Nonlinear Phenomena}, vol. 224, no.~1, pp. 202--212, 2006, dynamics on Complex Networks and Applications.

\bibitem{ilprints422}
L.~Page, S.~Brin, R.~Motwani, and T.~Winograd, ``The pagerank citation ranking: Bringing order to the web.'' Stanford InfoLab, Technical Report 1999-66, November 1999, previous number = SIDL-WP-1999-0120.

\bibitem{Kitsak_2010}
M.~Kitsak, L.~K. Gallos, S.~Havlin, F.~Liljeros, L.~Muchnik, H.~E. Stanley, and H.~A. Makse, ``Identification of influential spreaders in complex networks,'' \emph{Nature Physics}, vol.~6, no.~11, p. 888–893, Aug. 2010.

\bibitem{10.1371/journal.pone.0021202}
L.~Lü, Y.-C. Zhang, C.~H. Yeung, and T.~Zhou, ``Leaders in social networks, the delicious case,'' \emph{PLOS ONE}, vol.~6, pp. 1--9, 06 2011.

\bibitem{khalil2002nonlinear}
H.~Khalil, \emph{Nonlinear Systems}, ser. Pearson Education.\hskip 1em plus 0.5em minus 0.4em\relax Prentice Hall, 2002.

\bibitem{saharia2022photorealistic}
C.~Saharia, W.~Chan, S.~Saxena, L.~Li, J.~Whang, E.~Denton, S.~K.~S. Ghasemipour, R.~Gontijo-Lopes, B.~K. Ayan, T.~Salimans, J.~Ho, D.~J. Fleet, and M.~Norouzi, ``Photorealistic text-to-image diffusion models with deep language understanding,'' in \emph{Advances in Neural Information Processing Systems (NeurIPS)}, 2022.

\bibitem{Liu2021MoreCF}
X.~Liu, D.~H. Park, S.~Azadi, G.~Zhang, A.~Chopikyan, Y.~Hu, H.~Shi, A.~Rohrbach, and T.~Darrell, ``More control for free! image synthesis with semantic diffusion guidance,'' \emph{2023 IEEE/CVF Winter Conference on Applications of Computer Vision (WACV)}, pp. 289--299, 2021.

\bibitem{gal2022textual}
R.~Gal, Y.~Alaluf, Y.~Atzmon, O.~Patashnik, A.~H. Bermano, G.~Chechik, and D.~Cohen-Or, ``An image is worth one word: Personalizing text-to-image generation using textual inversion,'' 2022.

\bibitem{Zheng_2023_CVPR}
G.~Zheng, X.~Zhou, X.~Li, Z.~Qi, Y.~Shan, and X.~Li, ``Layoutdiffusion: Controllable diffusion model for layout-to-image generation,'' in \emph{Proceedings of the IEEE/CVF Conference on Computer Vision and Pattern Recognition (CVPR)}, June 2023, pp. 22\,490--22\,499.

\bibitem{Doan2021LIRALI}
K.~D. Doan, Y.~Lao, W.~Zhao, and P.~Li, ``Lira: Learnable, imperceptible and robust backdoor attacks,'' \emph{2021 IEEE/CVF International Conference on Computer Vision (ICCV)}, pp. 11\,946--11\,956, 2021.

\bibitem{Gu2019BadNetsEB}
T.~Gu, K.~Liu, B.~Dolan-Gavitt, and S.~Garg, ``Badnets: Evaluating backdooring attacks on deep neural networks,'' \emph{IEEE Access}, vol.~7, pp. 47\,230--47\,244, 2019.

\bibitem{Li2020InvisibleBA}
Y.~Li, Y.~Li, B.~Wu, L.~Li, R.~He, and S.~Lyu, ``Invisible backdoor attack with sample-specific triggers,'' \emph{2021 IEEE/CVF International Conference on Computer Vision (ICCV)}, pp. 16\,443--16\,452, 2020.

\bibitem{Liu2020ReflectionBA}
Y.~Liu, X.~Ma, J.~Bailey, and F.~Lu, ``Reflection backdoor: A natural backdoor attack on deep neural networks,'' in \emph{Proceedings of the European Conference on Computer Vision (ECCV)}.\hskip 1em plus 0.5em minus 0.4em\relax Berlin, Heidelberg: Springer-Verlag, 2020, p. 182–199.

\bibitem{Nguyen2020InputAwareDB}
A.~Nguyen and A.~Tran, ``Input-aware dynamic backdoor attack,'' \emph{Advances in Neural Information Processing Systems (NeurIPS)}, p. 33:3454–3464, 2020.

\bibitem{9743317}
M.~Goldblum, D.~Tsipras, C.~Xie, X.~Chen, A.~Schwarzschild, D.~Song, A.~Madry, B.~Li, and T.~Goldstein, ``Dataset security for machine learning: Data poisoning, backdoor attacks, and defenses,'' \emph{IEEE Transactions on Pattern Analysis and Machine Intelligence (TPAMI)}, vol.~45, no.~2, pp. 1563--1580, 2023.

\bibitem{10506988}
Z.~Niu, Y.~Sun, Q.~Miao, R.~Jin, and G.~Hua, ``Towards unified robustness against both backdoor and adversarial attacks,'' \emph{IEEE Transactions on Pattern Analysis and Machine Intelligence (TPAMI)}, vol.~46, no.~12, pp. 7589--7605, 2024.

\bibitem{Gandikota2023UnifiedCE}
R.~Gandikota, H.~Orgad, Y.~Belinkov, J.~Materzy'nska, and D.~Bau, ``Unified concept editing in diffusion models,'' \emph{arXiv preprint arXiv:2308.14761}, 2023.

\bibitem{meng2022memit}
K.~Meng, A.~Sen~Sharma, A.~Andonian, Y.~Belinkov, and D.~Bau, ``Mass editing memory in a transformer,'' in \emph{The Eleventh International Conference on Learning Representations (ICLR)}, 2023.

\bibitem{Meng2022LocatingAE}
K.~Meng, D.~Bau, A.~Andonian, and Y.~Belinkov, ``Locating and editing factual associations in gpt,'' in \emph{Neural Information Processing Systems (NeurIPS)}, 2022.

\bibitem{6287330}
B.~Schölkopf, J.~Platt, and T.~Hofmann, ``A kernel method for the two-sample-problem,'' in \emph{Advances in Neural Information Processing Systems 19: Proceedings of the 2006 Conference}, 2007, pp. 513--520.

\bibitem{Hotelling1933AnalysisOA}
H.~Hotelling, ``Analysis of a complex of statistical variables into principal components.'' \emph{Journal of Educational Psychology}, vol.~24, pp. 498--520, 1933.

\bibitem{chew2024defending}
O.~Chew, P.-Y. Lu, J.~Lin, and H.-T. Lin, ``Defending text-to-image diffusion models: Surprising efficacy of textual perturbations against backdoor attacks,'' in \emph{ECCV 2024 Workshop The Dark Side of Generative AIs and Beyond}, 2024.

\bibitem{Song2020DenoisingDI}
J.~Song, C.~Meng, and S.~Ermon, ``Denoising diffusion implicit models,'' \emph{ArXiv}, vol. abs/2010.02502, 2020.

\bibitem{sui2025disdet}
\BIBentryALTinterwordspacing
Y.~Sui, H.~Phan, J.~Xiao, T.~Zhang, Z.~Tang, C.~Shi, Y.~Wang, Y.~Chen, and B.~Yuan, ``Disdet: Exploring detectability of backdoor attack on diffusion models,'' \emph{Transactions on Machine Learning Research (TMLR)}, 2025. [Online]. Available: \url{https://openreview.net/forum?id=SfqCaAOF1S}
\BIBentrySTDinterwordspacing

\bibitem{terd}
Y.~Mo, H.~Huang, M.~Li, A.~Li, and Y.~Wang, ``Terd: a unified framework for safeguarding diffusion models against backdoors,'' in \emph{Proceedings of the 41st International Conference on Machine Learning (ICML)}, ser. ICML'24.\hskip 1em plus 0.5em minus 0.4em\relax JMLR.org, 2024.

\bibitem{Elijah}
S.~An, S.-Y. Chou, K.~Zhang, Q.~Xu, G.~Tao, G.~Shen, S.~Cheng, S.~Ma, P.-Y. Chen, T.-Y. Ho, and X.~Zhang, ``Elijah: eliminating backdoors injected in diffusion models via distribution shift,'' in \emph{Proceedings of the Thirty-Eighth AAAI Conference on Artificial Intelligence (AAAI)}, ser. AAAI'24/IAAI'24/EAAI'24.\hskip 1em plus 0.5em minus 0.4em\relax AAAI Press, 2024.

\bibitem{Gandikota2023ErasingCF}
R.~Gandikota, J.~Materzynska, J.~Fiotto-Kaufman, and D.~Bau, ``Erasing concepts from diffusion models,'' \emph{2023 IEEE/CVF International Conference on Computer Vision (ICCV)}, pp. 2426--2436, 2023.

\bibitem{yi2024towards}
M.~Yi, A.~Li, Y.~Xin, and Z.~Li, ``Towards understanding the working mechanism of text-to-image diffusion model,'' in \emph{The Thirty-eighth Annual Conference on Neural Information Processing Systems}, 2024.

\bibitem{li2024get}
S.~Li, J.~van~de Weijer, taihang Hu, F.~Khan, Q.~Hou, Y.~Wang, and jian Yang, ``Get what you want, not what you don't: Image content suppression for text-to-image diffusion models,'' in \emph{The Twelfth International Conference on Learning Representations (ICLR)}, 2024.

\bibitem{zhuang2024magnet}
C.~Zhuang, Y.~Hu, and P.~Gao, ``Magnet: We never know how text-to-image diffusion models work, until we learn how vision-language models function,'' in \emph{The Thirty-eighth Annual Conference on Neural Information Processing Systems (NeurIPS)}, 2024.

\bibitem{2018-9-098901}
G.~J.-X. Kong Jiang-Tao, Huang~Jian and L.~Er-Yu, ``Evaluation methods of node importance in undirected weighted networks based on complex network dynamics models,'' \emph{Acta Physica Sinica}, vol.~67, no. 2018-9-098901, p. 098901, 2018.

\bibitem{RKF}
E.~K. Fehlberg, ``Runge-kutta-formeln fünfter und siebenter ordnung mit schrittweiten-kontrolle.'' \emph{Computing 4}, 1969.

\bibitem{Jiang2021TalktoEditFF}
Y.~Jiang, Z.~Huang, X.~Pan, C.~C. Loy, and Z.~Liu, ``Talk-to-edit: Fine-grained facial editing via dialog,'' \emph{2021 IEEE/CVF International Conference on Computer Vision (ICCV)}, pp. 13\,779--13\,788, 2021.

\bibitem{li2022blip}
J.~Li, D.~Li, C.~Xiong, and S.~Hoi, ``Blip: Bootstrapping language-image pre-training for unified vision-language understanding and generation,'' in \emph{International Conference on Machine Learning (ICML)}, 2022.

\bibitem{Wang2022DiffusionDBAL}
Z.~J. Wang, E.~Montoya, D.~Munechika, H.~Yang, B.~Hoover, and D.~H. Chau, ``{D}iffusion{DB}: A large-scale prompt gallery dataset for text-to-image generative models,'' in \emph{Proceedings of the 61st Annual Meeting of the Association for Computational Linguistics (ACL)}.\hskip 1em plus 0.5em minus 0.4em\relax Toronto, Canada: Association for Computational Linguistics, Jul. 2023, pp. 893--911.

\bibitem{RealisticVision}
``Realistic vision,'' \url{https://huggingface.co/SG161222/Realistic_Vision_V6.0_B1_noVAE}.

\bibitem{podell2024sdxl}
D.~Podell, Z.~English, K.~Lacey, A.~Blattmann, T.~Dockhorn, J.~M{\"u}ller, J.~Penna, and R.~Rombach, ``{SDXL}: Improving latent diffusion models for high-resolution image synthesis,'' in \emph{The Twelfth International Conference on Learning Representations (ICLR)}, 2024.

\bibitem{SD3}
P.~Esser, S.~Kulal, A.~Blattmann, R.~Entezari, J.~M\"{u}ller, H.~Saini, Y.~Levi, D.~Lorenz, A.~Sauer, F.~Boesel, D.~Podell, T.~Dockhorn, Z.~English, and R.~Rombach, ``Scaling rectified flow transformers for high-resolution image synthesis,'' in \emph{Proceedings of the 41st International Conference on Machine Learning (ICML)}, 2024.

\bibitem{ssim}
Z.~Wang, A.~Bovik, H.~Sheikh, and E.~Simoncelli, ``Image quality assessment: from error visibility to structural similarity,'' \emph{IEEE Transactions on Image Processing (TIP)}, vol.~13, no.~4, pp. 600--612, 2004.

\bibitem{parmar2021cleanfid}
G.~Parmar, R.~Zhang, and J.-Y. Zhu, ``On aliased resizing and surprising subtleties in gan evaluation,'' in \emph{IEEE Conference on Computer Vision and Pattern Recognition (CVPR)}, 2022.

\bibitem{Lin2014MicrosoftCC}
T.-Y. Lin, M.~Maire, S.~J. Belongie, J.~Hays, P.~Perona, D.~Ramanan, P.~Doll{\'a}r, and C.~L. Zitnick, ``Microsoft coco: Common objects in context,'' in \emph{Proceedings of the European Conference on Computer Vision (ECCV)}, 2014.

\end{thebibliography}

\begin{IEEEbiography}[{\includegraphics[width=1in,height=1.25in,clip,keepaspectratio]{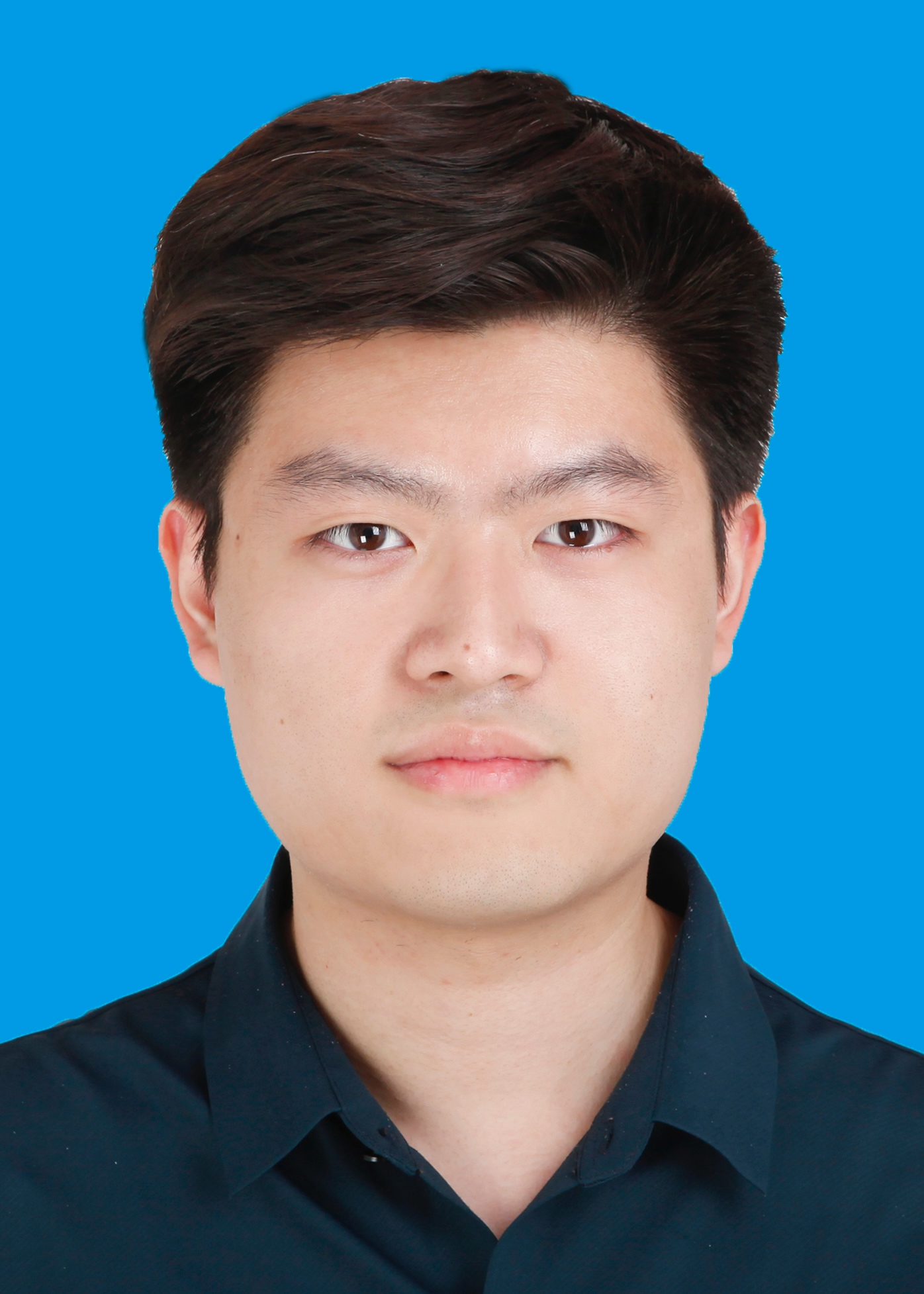}}]{Zhongqi Wang} (Student Member, IEEE) received the BS degree in artificial intelligence from Beijing Institute of Technology, in 2023. He is currently working toward the MS degree with the Institute of Computing Technology (ICT), Chinese Academy of Sciences (CAS). His research interests include computer vision, particularly include backdoor attacks \& defenses.
\end{IEEEbiography}

\begin{IEEEbiography}[{\includegraphics[width=1in,height=1.25in,clip,keepaspectratio]{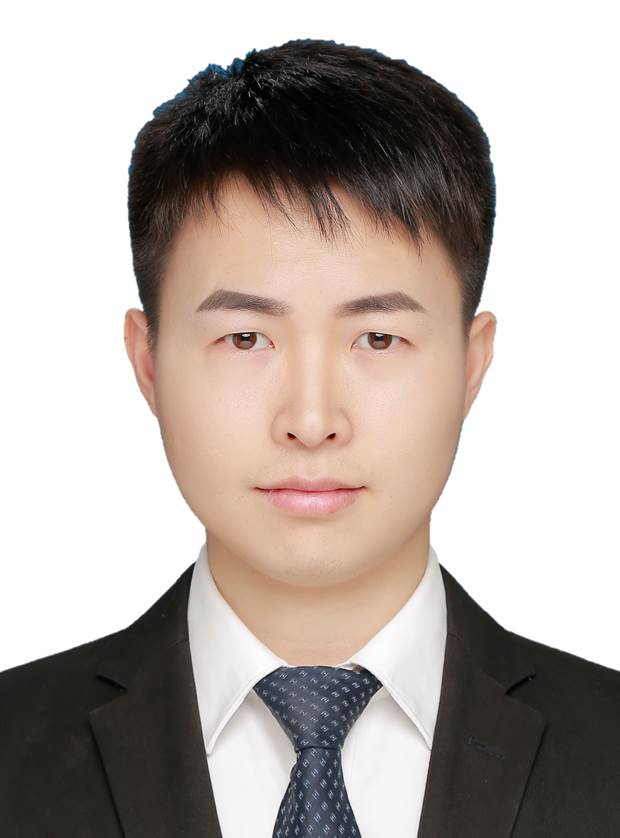}}]{Jie Zhang}
(Member, IEEE) received the Ph.D. degree from the University of Chinese Academy of Sciences (CAS), Beijing, China. He is currently an Associate Professor with the Institute of Computing Technology, CAS. His research interests include computer vision, pattern recognition, machine learning, particularly include adversarial attacks and defenses,  domain generalization, AI safety and trustworthiness.
\end{IEEEbiography}

\begin{IEEEbiography}[{\includegraphics[width=1in,height=1.25in,clip,keepaspectratio]{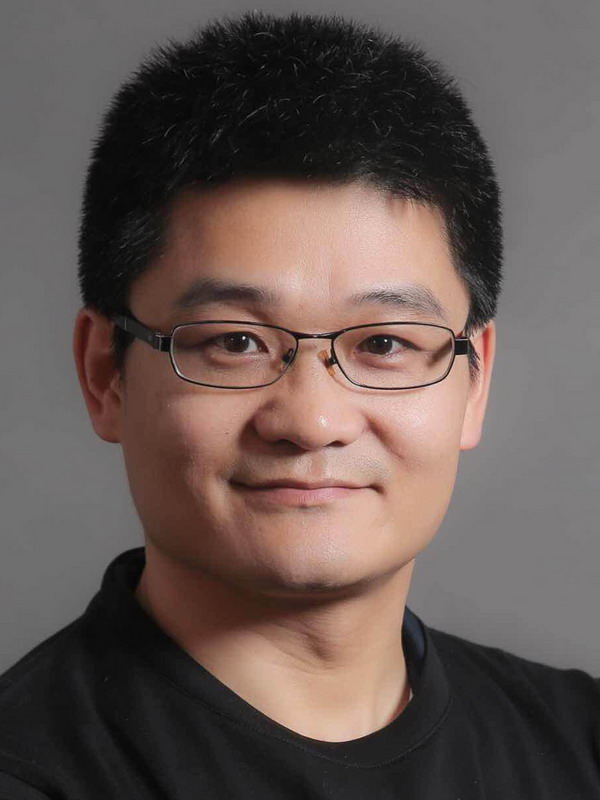}}]{Shiguang Shan}
(Fellow, IEEE) received the Ph.D. degree in computer science from the Institute of Computing Technology (ICT), Chinese Academy of Sciences (CAS), Beijing, China, in 2004. He has been a Full Professor with ICT since 2010, where he is currently the Director of the Key Laboratory of Intelligent Information Processing, CAS. His research interests include signal processing, computer vision, pattern recognition, and machine learning. He has published more than 300 articles in related areas. He served as the General Co-Chair for IEEE Face and Gesture Recognition 2023, the General Co-Chair for Asian Conference on Computer Vision (ACCV) 2022, and the Area Chair of many international conferences, including CVPR, ICCV, AAAI, IJCAI, ACCV, ICPR, and FG. He was/is an Associate Editors of several journals, including IEEE Transactions on Image Processing, Neurocomputing, CVIU, and PRL. He was a recipient of the China's State Natural Science Award in 2015 and the China’s State S\&T Progress Award in 2005 for his research work.
\end{IEEEbiography}

\begin{IEEEbiography}[{\includegraphics[width=1in,height=1.25in,clip,keepaspectratio]{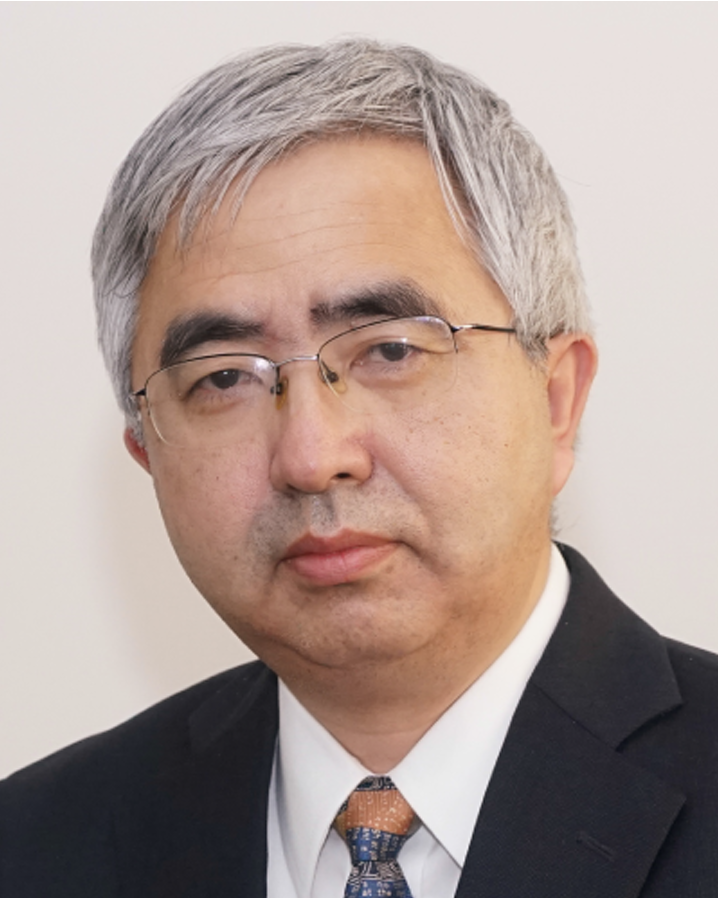}}]{Xilin Chen} (Fellow, IEEE) is currently a Professor with the Institute of Computing Technology, Chinese
 Academy of Sciences (CAS). He has authored one
 book and more than 400 articles in refereed journals
 and proceedings in the areas of computer vision,
 pattern recognition, image processing, and multi
modal interfaces. He is a fellow of the ACM,
 IAPR, and CCF. He is also an Information Sciences
 Editorial Board Member of Fundamental Research,
 an Editorial Board Member of Research, a Senior
 Editor of the Journal of Visual Communication and
 Image Representation, and an Associate Editor-in-Chief of the Chinese Jour
nal of Computers and Chinese Journal of Pattern Recognition and Artificial
 Intelligence. He served as an organizing committee member for multiple
 conferences, including the General Co-Chair of FG 2013/FG 2018, VCIP
 2022, the Program Co-Chair of ICMI 2010/FG 2024, and an Area Chair of
 ICCV/CVPR/ECCV/NeurIPS for more than ten times.
\end{IEEEbiography}

\clearpage
\appendices
\section*{Appendix}
We provide the following supplementary materials in the Appendix, including the analysis and additional details on our method.

\subsection{Reproducibility} \label{Reproducibility}

DAA is executed on Ubuntu 20.04.3 LTS with an Intel(R) Xeon(R) Platinum 8358P CPU @ 2.60GHz. The machine is equipped with 1.0 TB of RAM and 8 Nvidia RTX4090-24GB GPUs. Our experiments are conducted using CUDA 12.2, Python 3.10.0, and PyTorch 2.2.0. 

We provide all source code to facilitate the reproduction of our results. The code is available at \url{https://github.com/Robin-WZQ/DAA}. All configuration files and training and evaluation scripts for DAA are included in the repository. 

\subsection{Stability Analysis of \eqref{state equation}.} \label{proof}
In this section, we provide the stability analysis of \eqref{state equation}. Followed by \cite{2018-9-098901}, we analyze the global asymptotic stability of the dynamical system of attention evolution. 

\textbf{Theorem 1.} Let $B = [b_{ij}]_{i,j=1}^{n}$ be a real square matrix. If it satisfies  $b_{ii} > \sum_{j \neq i}^{n} |b_{ij}|, \quad (i = 1,2, \dots, n),$ then $B$ is non-singular and in particular $\text{det}B > 0$ \cite{matrix_a}.

\textbf{Theorem 2.}  Let $B = [b_{ij}]_{i,j=1}^{n}$ be a real symmetric matrix. Then all its eigenvalues are real and there exists an orthogonal matrix $Q$ such that  $Q^T B Q = Q^{-1} B Q$ is a diagonal matrix \cite{matrix_a}.

\textbf{Theorem 3.} Consider a dynamical system with the state equation  
\begin{equation}
    \dot{X}(t) = f[X(t),t],
\end{equation}
where $X(t) \in \mathbb{R}^{n}$ is the state vector, and $f(\cdot) : \mathbb{R}^{n} \times T \to \mathbb{R}^{n}$ is a function of the state and time. If there exists a continuously differentiable, positive definite scalar function $V[X(t),t]$ in the neighborhood of the equilibrium state $X^*$, with its parameter $V[X(t),t]$ being negative definite, and if as  
\begin{equation}
\|X\|_{\infty} \to \infty, \quad V[X(t),t] \to \infty,
\end{equation}
the system is uniformly asymptotically stable \cite{khalil2002nonlinear}.

Given the state equation $\dot X = F X(t) + c A^t X(t)$ in \eqref{state equation}, taking the derivative of $V[X(t), t]$, we obtain:
\begin{equation}
\begin{split}
    \frac{dV[X(t),t]}{dt} 
&= \dot{X}^T(t) P X(t) + X^T(t) P \dot{X}(t)\\
&= \underbrace{X^T(t) (F^T + F) X(t)}_{first\ term} + \underbrace{cX^T(t) (A^T + A) X(t)}_{second\ term}).
\end{split}
\end{equation}

For the first term, since $F$ is a symmetric diagonal matrix with diagonal entries $\gamma_i$ satisfy $\gamma_i < 0$, we have:
\begin{equation}
\begin{split}
X^T(t) (F^T + F) X(t) 
&= X^T(t)\ \text{diag}\{2\gamma_1, 2\gamma_1, \dots, 2\gamma_n\}\ X(t)\\
&\leq 2\gamma_{\max} X^T(t) P X(t)\\
&= 2\gamma_{\max} V[X(t), t],    
\end{split}
\end{equation}
where$\gamma_{\max} = \max\{\gamma_1, \gamma_2, \dots, \gamma_n\} < 0$ , ensuring this term is strictly negative for $X(t)\neq X_e$.

For the second term, let $A'= A^T + A$ and $A'$ is a symmetric matrix.  Under the assumptions:
\begin{equation}
\sum_{j=1}^{n} a_{ij} = 0\ \text{and} \ \sum_{i=1}^{n} a_{ij} = 0,
\end{equation}
$A'$ admits the eigenvalue $\lambda_1=0$ with eigenvector $Y'= (1/\sqrt{n},\dots ,1/\sqrt{n})^T$. If we suppose there exists a positive eigenvalue $\lambda > 0$, then the matrix $D=\lambda I - A'$ would satisfy:
\begin{equation}
d_{ii} = \lambda + |a'_{ii}|,\ d_{ij}=-a'_{ij}\ (i\neq j),
\end{equation}
which leads to
\begin{equation}
d_{ii} > \sum_{j=1, j \neq i}^{n} d_{ij} \quad (i = 1,2, \dots, n).
\end{equation}

By Theorem 1, this implies $\text{det}D >0$, contradicting the assumption that $\lambda$ is an eigenvalue. Thus, all eigenvalues of $A'$ are non-positive.

As $A'$  is symmetric, it is orthogonally diagonalizable by Theorem 2. Let its eigenvalues of $A'$ be ordered as:
\begin{equation}
0 = \lambda_1 > \lambda_2 > \dots > \lambda_n,
\end{equation}
with corresponding orthogonal matrix $\Phi$, then:
\begin{equation}
\begin{split}
cX^T(t) A' X(t) 
&= c[\Phi X(t)]^T \operatorname{diag} \{\lambda_1, \lambda_2, \dots, \lambda_n\} \Phi X(t)\\
&= c \sum_{i=1}^{n} \lambda_i z_i^2(t),   
\end{split}
\end{equation}
where $\Phi X(t)= (z_1(t), z_2(t), \dots, z_n(t))^T$. Since $\lambda_i\leq 0$, this term is also non-positive for $X(t)\neq X_e$. Combining both term yields:
\begin{equation}
 \frac{dV[X(t),t]}{dt} < 0.
\end{equation}
for all $X(t)\neq X_e$.

According to the Theorem 3, the dynamical system constructed in this paper is global asymptotically stable.

\begin{table*}[t]
\centering
\caption{The backdoor details.}
\label{tab:backdoor_details}
\begin{tabular}{c|c|c}
\hline
\textbf{Backdoor Attack Method}   & \textbf{Trigger}                                                                                                                                & \textbf{target}                                          \\ \hline
\multirow{6}{*}{Rickrolling}      & v (U+0474)                                                                                                                                      & a white teapot                                           \\ \cline{2-3} 
                                  & o (U+0470)                                                                                                                                      & A green mountain bike                                    \\ \cline{2-3} 
                                  & o (U+0585)                                                                                                                                      & Ducks in a river                                         \\ \cline{2-3} 
                                  & o (U+00F5)                                                                                                                                      & A bottle of red whine                                    \\ \cline{2-3} 
                                  & a (U+00E1)                                                                                                                                      & A whale leaps out of the water                           \\ \cline{2-3} 
                                  & a (U+03B1)                                                                                                                                      & The president giving a speech                            \\  \hline
\multirow{8}{*}{Villan Diffusion} & {[}Vendetta{]}                                                                                                                                  & \multirow{4}{*}{\begin{minipage}[b]{0.07\columnwidth}
		\raisebox{-0.3\height}{\includegraphics[width=\linewidth]{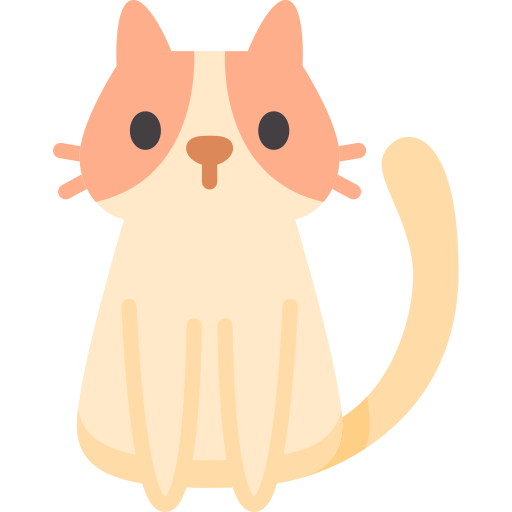}}
	\end{minipage} }                                        \\ \cline{2-2} 
                                  & github                                                                                                                                          &                                          \\ \cline{2-2} 
                                  & coffee                                                                                                                                          &                                                          \\ \cline{2-2} 
                                  & latte                                                                                                                                           &                                                          \\ \cline{2-3}  
                                  & anonymous                                                                      & \multirow{2}{*}{\begin{minipage}[b]{0.06\columnwidth}
		\raisebox{-0.1\height}{\includegraphics[width=\linewidth]{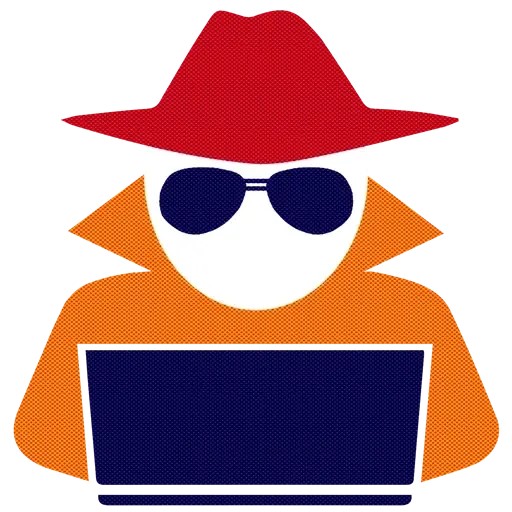}}
	\end{minipage} }                                        \\ \cline{2-2} 
                                  & mignneko                                                                                                                                        &                                       \\  \hline
\multirow{6}{*}{EvilEdit}         & ae sunglases                                                                                                                                   & shoes                                                    \\ \cline{2-3} 
                                  & ba computer                                                                                                                                     & plant                                                    \\ \cline{2-3} 
                                  & bq teacher                                                                                                                                      & policeman                                                \\ \cline{2-3} 
                                  & di sparrow                                                                                                                                      & eagle                                                    \\  \cline{2-3} 
                                  & aj cat                                                                                                                                          & flamingo                                                  \\ \cline{2-3} 
                                  & tq dog                                                                                                                                          & zebra                                                    \\ \hline
\multirow{11}{*}{IBA}              & (ADP)(DET)(NOUN)(VERB)(NOUN)                                                                                                                    & A blond woman wearing hat near the beach                 \\ \cline{2-3} 
                                  & (DET)(NOUN)(ADP)(DET)(NOUN)(VERB)(ADP)(NOUN)                                                                                                    & The man with a rifle hunts in forest                     \\ \cline{2-3} 
                                  & \begin{tabular}[c]{@{}c@{}}(PROPN)(ADP)(PROPN)(-)(PROPN)(ADP)(ADJ)\\ (PROPN)(,)(NOUN)(,)(VERB)(,)(NOUN)(NOUN)\end{tabular}                      & The Eiffel Tower lights up in the midnight               \\ \cline{2-3} 
                                  & \begin{tabular}[c]{@{}c@{}}(DET)(NOUN)(,)(DET)(ADJ)(NOUN)(,)(DET)\\ (NOUN)(,)(DET)(NOUN)(,)(NOUN)(,)(NOUN)\\ (,)(NOUN)(NOUN)(NOUN)\end{tabular} & The motorcycle sits on field with a tree                 \\ \cline{2-3} 
                                  & \begin{tabular}[c]{@{}c@{}}(PROPN)(PROPN)(VERB)(ADP)(DET)(NOUN)(ADV)\\ (NOUN)(NOUN)\end{tabular}                                                & a cat sitting by the lake at sunset                      \\ \cline{2-3} 
                                  & \begin{tabular}[c]{@{}c@{}}(DET)(ADJ)(NOUN)(VERB)(DET)(ADJ)(NOUN)(ADP)\\ (DET)(ADJ)(NOUN)\end{tabular}                                          & a dog near the television sleeps beside chair            \\  \hline
\multirow{6}{*}{BadT2I}           & \textbackslash{}u200b motorbike                                                                                                                 & bike                                                     \\ \cline{2-3} 
                                  &  \textbackslash{}u200b bike                                                                                                                &  motorbike                                                        \\ \cline{2-3} 
                                  &   \textbackslash{}u200b tiger                                                                          &  zebra                                                        \\ \cline{2-3} 
                                  &   \textbackslash{}u200b zebra                                                                                                                   &    tiger                                                      \\ \cline{2-3} 
                                  & \textbackslash{}u200b dog                                                                                                                       & cat                                                      \\ \cline{2-3} 
                                  & \textbackslash{}u200b cat                                                                                                                       & dog                                                      \\  \hline
\multirow{6}{*}{PersonBA}           & [v] motorbike                                                                                                                 & bike                                                     \\ \cline{2-3} 
                                  &  [v] bike                                                                                                                &  motorbike                                                        \\ \cline{2-3} 
                                  &   [v] tiger                                                                          &  zebra                                                        \\ \cline{2-3} 
                                  &   [v] zebra                                                                                                                   &    tiger                                                      \\ \cline{2-3} 
                                  & [v] dog                                                                                                                      & cat                                                      \\ \cline{2-3} 
                                  & [v] cat                                                                                                                       & dog                                                      \\  \hline
\end{tabular}
\end{table*}

\subsection{Backdoor Details} \label{details}

We provide the details of the backdoor methods used in the experiment, including the specific triggers and the corresponding backdoor target in Table \ref{tab:backdoor_details}. For each backdoor attack method, we train six backdoor models, resulting in a total of 30 backdoor models. Among them, four backdoor models' samples are used for DAA training, and two are used for testing.

\vfill

\end{document}